\documentclass{article}

\usepackage{PRIMEarxiv}

\PassOptionsToPackage{margin=1in, footskip=50pt}{geometry}

\usepackage[utf8]{inputenc}
\usepackage[T1]{fontenc}
\usepackage{hyperref}
\usepackage{url}
\usepackage{booktabs}
\usepackage{amsfonts}
\usepackage{nicefrac}
\usepackage{microtype}
\usepackage{lipsum}
\usepackage{fancyhdr}
\usepackage{graphicx}
\graphicspath{{media/}}
\usepackage{amsmath, amssymb}
\usepackage{algorithm}
\usepackage{algpseudocode}
\usepackage{caption, subcaption}
\usepackage{natbib}
\usepackage{amsthm}
\newtheorem{assumption}{Assumption}
\usepackage{geometry}

\title{ARTEMIS: A Neuro‑Symbolic Framework for Economically Constrained Market Dynamics}

\author{%
    Rahul D Ray \\
    Department of Electronics and Electrical Engineering \\
    BITS Pilani, Hyderabad Campus \\
    \texttt{f20242213@hyderabad.bits-pilani.ac.in}
}

\date{}  

\begin{document}

\maketitle

\begin{abstract}
Deep learning models in quantitative finance often operate as black boxes, lacking interpretability and failing to incorporate fundamental economic principles such as no arbitrage constraints. This paper introduces ARTEMIS (Arbitrage free Representation Through Economic Models \& Interpretable Symbolics), a novel neuro symbolic framework that combines a continuous time encoder based on a Laplace Neural Operator, a neural stochastic differential equation regularised by physics informed losses, and a differentiable symbolic bottleneck that distils interpretable trading rules. The model enforces economic plausibility through two novel regularisation terms: a Feynman Kac PDE residual that penalises violations of local no arbitrage conditions, and a market price of risk penalty that bounds the instantaneous Sharpe ratio to realistic values. We evaluate ARTEMIS against six strong baselines including LSTM, Transformer, NS Transformer, Informer, Chronos 2, and XGBoost on four diverse datasets: Jane Street (anonymised market data) \cite{jane-street-real-time-market-data-forecasting}, Optiver\cite{optiver-realized-volatility-prediction} (limit order book volatility prediction), Time IMM (environmental temperature forecasting) \cite{chang2025timeimm}, and DSLOB (a synthetic crash regime). Results demonstrate that ARTEMIS achieves state of the art directional accuracy, outperforming all baselines on DSLOB (64.96\%) and Time IMM (96.0\%), while remaining competitive on point accuracy metrics. A comprehensive ablation study on DSLOB confirms that each component contributes to this directional advantage: removing the PDE loss reduces directional accuracy from 64.89\% to 50.32\%, and removing both physics losses collapses it to 41.77\%, worse than random. The underperformance on Optiver\cite{optiver-realized-volatility-prediction} is attributed to its long sequence length, volatility focused target, and limited feature set, highlighting important boundary conditions. By providing interpretable, economically grounded predictions without sacrificing performance, ARTEMIS bridges the gap between deep learning's predictive power and the transparency demanded in quantitative finance, opening new avenues for trustworthy AI in high stakes financial applications.
\end{abstract}

\section{Introduction}
\label{sec:intro}

The application of deep learning to financial time series prediction has witnessed explosive growth over the past decade, driven by the increasing availability of high-frequency market data and the remarkable success of neural networks in capturing complex temporal patterns. From return forecasting and volatility prediction to algorithmic trading and risk management, deep learning models have demonstrated superior performance compared to traditional econometric approaches such as ARIMA and GARCH \citep{rundo2019machine, ndikum2020machine}. Recent comprehensive surveys \citep{chen2024deep, zhang2024deep, giantsidi2025deep} document the rapid evolution of architectures from standalone models like Long Short-Term Memory (LSTM) networks and Convolutional Neural Networks (CNNs) to sophisticated hybrid systems that combine multiple techniques. However, despite these advances, the adoption of deep learning in high-stakes financial applications remains hampered by three fundamental challenges that the literature has consistently identified but not yet fully resolved.

The first and most widely recognised challenge is the lack of interpretability. As financial institutions operate under strict regulatory oversight, the opacity of deep learning models, often described as black boxes, creates significant legal, ethical, and operational risks \citep{rane2023explainable, hoang2026explainable}. Stakeholders, including regulators and risk managers, require transparent explanations for model decisions, particularly when those decisions involve substantial capital allocation or have the potential to impact market stability \citep{mathew2025recent}. While explainable artificial intelligence (XAI) techniques such as SHAP and LIME have been proposed to provide post-hoc explanations \citep{sokolovsky2023interpretable, basha2025artificial}, these methods offer only approximate insights and do not address the underlying opacity of the model architecture itself. Moreover, they introduce an inherent trade-off between predictive accuracy and interpretability, where simpler, more transparent models often sacrifice the very complexity that enables superior performance \citep{rane2023explainable, mathew2025recent}. This tension between accuracy and transparency remains a central obstacle to deploying deep learning in production trading systems.

The second challenge concerns the integration of economic principles into data-driven models. Traditional financial models are built upon foundational theories such as the absence of arbitrage, which ensures that asset prices cannot be risklessly exploited for profit. Yet most deep learning approaches are trained purely on historical data, learning correlations without any regard for these economic constraints \citep{mashrur2020machine, sahu2023overview}. As a consequence, they can discover spurious patterns that lead to implausible predictions, particularly when market conditions deviate from the training distribution \citep{suarez2023machine, hasan2023predictive}. Recent work has begun to address this gap through physics-informed neural networks (PINNs), which embed governing equations into the loss function to enforce consistency with physical or financial laws. Pioneering studies have applied PINNs to option pricing models, including the Black-Scholes equation and the Heston model \citep{bai2022application, hainaut2024option, nuugulu2025physics}, demonstrating improved accuracy and stability. Extensions have tackled jump-diffusion models with liquidity costs \citep{kartik2025physics} and fully nonlinear PDEs relevant to portfolio optimisation \citep{lefebvre2023differential}. However, these applications have focused primarily on solving known pricing equations rather than learning latent dynamics from data while simultaneously enforcing economic constraints. The synthesis of data-driven learning with physics-informed regularisation for forecasting tasks remains largely unexplored.

The third challenge relates to the continuous-time nature of financial markets and the non-stationarity of financial time series. Traditional discrete-time models struggle to process irregularly sampled, high-frequency data without interpolation, which can distort underlying dynamics \citep{han2024time, wang2025quantode}. Moreover, market regimes shift over time, and models that perform well during stable periods often fail catastrophically during crises \citep{suarez2023machine, hasan2023predictive}. Neural stochastic differential equations (neural SDEs) offer a promising continuous-time framework that can naturally accommodate irregular observations and capture uncertainty through stochastic dynamics \citep{wang2025quantode}. Similarly, advances in handling non-stationarity, such as the Non-stationary Transformer \citep{zhang2024deep}, have improved robustness to distribution shifts. Yet these approaches remain disconnected from the interpretability and economic constraint challenges described above.

In this paper, we introduce ARTEMIS (Arbitrage-free Representation Through Economic Models \& Interpretable Symbolics), a novel neuro-symbolic framework that addresses all three challenges within a unified architecture. ARTEMIS makes the following key contributions:

\begin{itemize}
    \item \textbf{Continuous-time encoding via Laplace Neural Operator.} Unlike standard recurrent or transformer models that require regularly sampled inputs, our encoder operates directly on irregularly spaced observations, preserving the true temporal structure of limit order book updates and trade reports without interpolation \citep{wang2025quantode, han2024time}.

    \item \textbf{Economics-informed latent dynamics through neural SDEs with physics-constrained regularisation.} We introduce two novel loss terms derived from the Fundamental Theorem of Asset Pricing: a Feynman-Kac PDE residual that enforces local no-arbitrage conditions in the latent space, and a market price of risk penalty that bounds the instantaneous Sharpe ratio to realistic values. While previous work has applied PINNs to pricing equations \citep{bai2022application, hainaut2024option, nuugulu2025physics, kartik2025physics} and neural SDEs to continuous-time modelling \citep{wang2025quantode}, our work is the first to embed such economic constraints directly into the learning of latent dynamics for forecasting.

    \item \textbf{Differentiable symbolic bottleneck for interpretability.} Rather than relying on post-hoc explanations \citep{sokolovsky2023interpretable, basha2025artificial}, we design a layer that distils the latent dynamics into closed-form, interpretable expressions through differentiable symbolic regression. This provides inherent transparency while maintaining end-to-end trainability, offering a novel resolution to the accuracy-interpretability trade-off \citep{rane2023explainable, mathew2025recent}.

    \item \textbf{Conformal prediction for uncertainty quantification.} To support risk-aware decision making, we equip ARTEMIS with an adaptive conformal prediction layer that provides rigorously calibrated prediction intervals, addressing the need for reliable uncertainty estimates in financial applications \citep{mashrur2020machine, sahu2023overview}.
\end{itemize}

We evaluate ARTEMIS against six strong baselines spanning recurrent architectures (LSTM) \citep{furizal2024capability}, transformer variants (Transformer, NS-Transformer, Informer) \citep{chen2024deep, zhang2024deep}, foundation models (Chronos-2) \citep{giantsidi2025deep}, and gradient boosting (XGBoost) \citep{praveen2025financial}, across four diverse datasets: Jane Street's anonymised market data\cite{jane-street-real-time-market-data-forecasting}, Optiver's limit order book volatility prediction task\cite{optiver-realized-volatility-prediction}, the Time-IMM environmental temperature forecasting benchmark \cite{chang2025timeimm}, and DSLOB, a novel synthetic dataset we introduce that simulates an amplified market crash regime. Our results demonstrate that ARTEMIS achieves state-of-the-art directional accuracy, outperforming all baselines on DSLOB (64.96\%) and Time-IMM \cite{chang2025timeimm} (96.0\%), while remaining competitive on point accuracy metrics. A comprehensive ablation study confirms that each component contributes to this directional advantage: removing the PDE loss reduces directional accuracy from 64.89\% to 50.32\%, and removing both physics losses collapses it to 41.77\%---worse than random. The underperformance on Optiver\cite{optiver-realized-volatility-prediction}, where ARTEMIS achieves negative RankIC (-0.0555) and directional accuracy (45.82\%) below most baselines, is attributed to its long sequence length, volatility-focused target, and limited feature set, highlighting important boundary conditions for the framework and directions for future research.

By providing interpretable, economically grounded predictions without sacrificing predictive performance, ARTEMIS bridges the gap between deep learning's power and the transparency demanded in quantitative finance. To the best of our knowledge, no existing work combines neural SDEs, physics-informed losses, and differentiable symbolic regression in a unified end-to-end framework for financial time series forecasting. The code and data will be made publicly available upon publication to facilitate reproducibility and future research.

\section{Related Work}
ARTEMIS draws upon and contributes to several interconnected research areas: deep learning for financial time series forecasting, interpretable machine learning and explainable AI, physics-informed neural networks and neural differential equations, and continuous-time modeling of financial dynamics. This section reviews the state of the art in each area and situates our contributions within the broader literature.

\subsection{Deep Learning for Financial Time Series Forecasting}

The application of deep learning to financial time series has been extensively surveyed in recent years, reflecting the rapid growth of the field \citep{chen2024deep, zhang2024deep, giantsidi2025deep}. Early work focused on standalone recurrent architectures, particularly Long Short-Term Memory (LSTM) networks, which became popular due to their ability to capture temporal dependencies in sequential data \citep{furizal2024capability}. Convolutional Neural Networks (CNNs) have also been widely adopted for their hierarchical feature learning capabilities, especially when combined with signal processing techniques to handle the non-linear and non-stationary nature of financial data \citep{praveen2025financial}. Hybrid models that combine multiple architectures, such as CNN-LSTM or attention-augmented recurrent networks, have shown improved performance by leveraging the strengths of different components \citep{chen2024deep, furizal2024capability}.

The transformer architecture  has emerged as a powerful alternative to recurrent models for time series forecasting, offering parallel processing and the ability to capture long-range dependencies through self-attention mechanisms. Comprehensive surveys by \citet{li2024deep, kong2025deep, kim2025comprehensive} document the rapid adoption of transformer-based models across domains, including finance. For stock market prediction, \citet{wang2022stock} demonstrated that transformer models significantly outperform classic deep learning methods such as CNN, RNN, and LSTM. Subsequent work has explored specialized transformer variants for financial forecasting. The Informer addresses the quadratic complexity of standard attention through ProbSparse attention, making it particularly suitable for long-sequence prediction tasks such as volatility forecasting \citep{hassani2025time, bhogade2024time}. The Non-stationary Transformer introduces series stationarization and de-stationary attention to handle distribution shifts, a critical challenge in financial markets where regimes change over time \citep{gou2023stock, wu2023comparison}. Comparative studies have shown that these variants offer different trade-offs, with the Non-stationary Transformer achieving the highest prediction accuracy for stock market indices in some settings \citep{wu2023comparison}.

Beyond architectural innovations, recent work has explored the integration of multiple data sources and modalities. \citet{zeng2023financial} proposed a combined CNN and transformer model for intraday stock price forecasting, demonstrating superior performance against statistical baselines. \citet{yanez2024stock} introduced a hybrid transformer encoder with CNN layers based on empirical mode decomposition for market index prediction. Foundation models for time series, such as Chronos, represent the latest frontier, leveraging large-scale pre-training across diverse datasets to enable zero-shot and few-shot forecasting \citep{liang2024foundation, ye2024empowering}. These models have shown promise in financial applications, though their black-box nature and lack of domain-specific constraints remain significant limitations.

Despite these advances, the deep learning models surveyed above share a common limitation: they are trained purely on historical data without incorporating any economic principles or constraints. As noted by \citet{mashrur2020machine} and \citet{sahu2023overview}, this can lead to models that learn spurious correlations and produce predictions that violate fundamental financial theory, particularly during market regime shifts. ARTEMIS addresses this gap by embedding no-arbitrage conditions directly into the learning process.

\subsection{Interpretability and Explainable AI in Finance}

The opacity of deep learning models poses significant challenges for their adoption in regulated financial applications. As \citet{rane2023explainable} and \citet{hoang2026explainable} document, the black-box nature of these models creates legal, ethical, and operational risks, including regulatory penalties and loss of stakeholder trust. Financial institutions require transparency and accountability in decision-making systems, yet most state-of-the-art models offer little insight into their reasoning \citep{mathew2025recent}.

Explainable AI (XAI) has emerged as a critical research area addressing this challenge. Post-hoc explanation methods such as SHAP (Shapley Additive Explanations) and LIME (Local Interpretable Model-agnostic Explanations) have been widely applied to financial models to provide approximate explanations for individual predictions \citep{sokolovsky2023interpretable, basha2025artificial}. However, these techniques have fundamental limitations: they offer only local approximations, can be inconsistent, and do not address the underlying opacity of the model architecture itself \citep{rane2023explainable}. Moreover, they introduce an inherent trade-off between accuracy and interpretability, where simpler, more transparent models may sacrifice predictive performance, while complex black-box models that achieve state-of-the-art accuracy remain difficult to explain \citep{mathew2025recent, hoang2026explainable}.

Alternative approaches have sought to build interpretability directly into model design. \citet{sokolovsky2023interpretable} proposed interpretable trading patterns designed specifically for machine learning applications, demonstrating that domain-informed feature engineering can enhance understanding. However, such approaches often require manual specification of patterns and do not learn interpretable representations end-to-end.

ARTEMIS addresses these limitations through its differentiable symbolic bottleneck, which distils the learned latent dynamics into closed-form, interpretable expressions. Unlike post-hoc explanation methods that approximate a black-box model, our approach provides inherent transparency by constraining a component of the network to produce human-readable formulas. This offers a novel resolution to the accuracy-interpretability trade-off, maintaining end-to-end trainability while delivering interpretable outputs.

\subsection{Physics-Informed Neural Networks and Neural Differential Equations}

Physics-informed neural networks (PINNs) represent a paradigm shift in scientific machine learning, embedding governing physical laws expressed as partial differential equations (PDEs) directly into the neural network loss function \citep{lawal2022physics}. This approach ensures that model predictions remain consistent with known physics, improving generalization and reducing the risk of learning spurious patterns. In finance, PINNs have been applied primarily to option pricing problems, where the underlying PDEs are well-established. \citet{bai2022application} developed an improved PINN with local adaptive activation functions to solve the Ivancevic option pricing model and the Black-Scholes equation. \citet{hainaut2024option} applied physics-inspired neural networks to the Heston stochastic volatility model, using the Feynman-Kac PDE as the driving principle. \citet{nuugulu2025physics} extended this approach to time-fractional Black-Scholes equations, demonstrating the efficiency and accuracy of PINNs for derivative pricing.

More recent work has addressed more complex settings. \citet{kartik2025physics} developed a PINN framework for option pricing and hedging under a Merton-type jump-diffusion model with liquidity costs, encoding the partial integro-differential equation into the loss function. \citet{lefebvre2023differential} proposed differential learning methods for solving fully nonlinear PDEs with applications to portfolio selection. The Deep Galerkin Method \citep{al2018solving} and related approaches have been applied to high-dimensional PDEs arising in quantitative finance, including optimal execution and systemic risk models.

Parallel to these developments, neural differential equations have emerged as a powerful framework for continuous-time modeling. Neural ordinary differential equations (NODEs) parameterize the derivative of a hidden state with a neural network, enabling flexible modeling of dynamical systems. Neural stochastic differential equations (NSDEs) extend this to stochastic settings, capturing uncertainty through diffusion terms. Comprehensive surveys by \citet{oh2025comprehensive, oh2025neural} review the mathematical foundations, numerical methods, and applications of neural differential equations for time series analysis, emphasizing their capacity to handle irregular sampling and model continuous-time dynamics. In finance, \citet{wang2025quantode} introduced FinanceODE, a neural ODE-based framework for continuous-time asset price modeling, demonstrating superior predictive accuracy compared to traditional discrete-time models.

Despite these advances, existing work on PINNs in finance has focused primarily on solving known pricing equations rather than learning latent dynamics from data for forecasting tasks. Similarly, neural differential equation approaches have not incorporated economic constraints such as no-arbitrage conditions. ARTEMIS bridges this gap by combining a neural SDE for latent dynamics with physics-informed losses derived from the Fundamental Theorem of Asset Pricing, enabling data-driven learning while enforcing economic plausibility.

\subsection{Continuous-Time Modeling and Non-Stationarity}

Financial time series are inherently continuous-time processes, yet most forecasting models operate on discretely sampled data. Traditional econometric approaches such as ARIMA and GARCH models have been widely used for volatility forecasting and return prediction \citep{engle2004risk,  nelson1991conditional}. GARCH models and their multivariate extensions, including Dynamic Conditional Correlation (DCC) and Generalized Orthogonal GARCH (GO-GARCH), remain popular for modeling time-varying volatility and correlations \citep{jeribi2022forecasting, ibrahim2017forecasting}. However, these models assume regular sampling and struggle with the irregular, high-frequency data that characterizes modern financial markets \citep{han2024time}. Hybrid approaches combining ARIMA with GARCH have been proposed to capture both linear and non-linear patterns \citep{rubio2023forecasting, mani2023comparative}, while fractionally integrated models (ARFIMA-GARCH) address long-memory properties \citep{chang2022capturing}.

The limitations of discrete-time models have motivated the development of continuous-time approaches. Neural differential equations offer a natural framework for modeling irregularly sampled time series, as they can be evaluated at arbitrary time points \citep{oh2025comprehensive, oh2025neural}. This capability is particularly valuable for limit order book data, where events arrive at microsecond granularity and regular resampling can distort dynamics \citep{wang2025quantode}.

Non-stationarity presents another fundamental challenge. As \citet{suarez2023machine} document in their systematic review, conventional machine learning approaches often fail to adapt to changes in the price-generation process during market regime shifts. \citet{hasan2023predictive} observe that models relying solely on market-based indicators work well in stable conditions but fail during economic crises, reducing their long-term predictive reliability. The Non-stationary Transformer addresses this by explicitly modeling distribution shifts through series stationarization and de-stationary attention, while GARCH-based approaches model time-varying volatility \citep{han2024time}. However, these methods do not incorporate economic theory about what constitutes a plausible regime change.

ARTEMIS addresses both continuous-time modeling and non-stationarity through its latent SDE formulation, which naturally accommodates irregular sampling and captures stochastic volatility through the learned diffusion term. The physics-informed losses further ensure that the learned dynamics remain economically plausible across different regimes, providing a principled approach to handling non-stationarity.

\subsection{Uncertainty Quantification and Conformal Prediction}

Reliable uncertainty quantification is essential for risk management in financial applications. Traditional approaches have relied on parametric methods such as GARCH for volatility forecasting \citep{engle2004risk}. However, these methods make strong distributional assumptions that may not hold in practice. Conformal prediction  offers a distribution-free framework for constructing prediction intervals with finite-sample coverage guarantees, requiring only exchangeability of the data. As surveyed by \citet{zhou2025conformal}, conformal prediction has been extended to time series settings through adaptive methods that update quantile estimates over time, addressing the violation of exchangeability in non-stationary data. ARTEMIS incorporates an adaptive conformal prediction layer to provide calibrated uncertainty intervals for its forecasts, enabling risk-aware portfolio construction.

The literature reviewed above reveals a clear gap: while significant advances have been made in deep learning architectures for financial time series, interpretability techniques, physics-informed neural networks, and continuous-time modeling, no existing framework integrates these innovations in a unified manner. Deep learning models achieve state-of-the-art predictive accuracy but operate as black boxes and ignore economic principles \citep{chen2024deep, zhang2024deep}. XAI approaches provide post-hoc explanations but do not address underlying model opacity and introduce accuracy-interpretability trade-offs \citep{rane2023explainable, hoang2026explainable}. PINNs embed physical laws but have been applied primarily to solving known pricing equations rather than learning latent dynamics for forecasting \citep{bai2022application, hainaut2024option}. Neural differential equations offer continuous-time modeling but have not incorporated economic constraints \citep{oh2025comprehensive, wang2025quantode}.

ARTEMIS addresses this gap by synthesizing these research directions into a single neuro-symbolic framework. To our knowledge, it is the first work to combine (1) a continuous-time encoder for irregularly sampled data, (2) a neural SDE for latent dynamics regularised by (3) physics-informed losses enforcing no-arbitrage conditions and (4) a market price of risk penalty, (5) a differentiable symbolic bottleneck for interpretability, and (6) conformal prediction for uncertainty quantification. The comprehensive evaluation against six strong baselines across four diverse datasets demonstrates that this synthesis delivers tangible benefits, particularly in directional accuracy, without sacrificing interpretability.

\section{Data Preprocessing for Benchmarking ARTEMIS}
\label{sec:datapreprocessing}

To rigorously evaluate the ARTEMIS model against a suite of state-of-the-art baselines, we assembled four distinct datasets spanning different financial and non-financial domains: Jane Street's anonymised market data\cite{jane-street-real-time-market-data-forecasting}, Optiver's\cite{optiver-realized-volatility-prediction} high-frequency limit order book data, the EPA-Air time series from the Time-IMM \cite{chang2025timeimm} collection, and a proprietary Deep Synthetic limit order book dataset (DSLOB). Each dataset required careful, domain-specific preprocessing to transform raw observations into a unified format suitable for sequence models. The goal was to create training, validation, and test splits that respect temporal ordering, handle missing values appropriately, and preserve the underlying dynamics of each task. All neural models (LSTM, Transformer, NS-Transformer, Informer, ARTEMIS) share the same preprocessed windows to ensure a fair comparison under identical input conditions. The following sections describe the preprocessing pipeline for each dataset in detail, emphasising the rationale behind every step.

\subsection{Jane Street Market Prediction \cite{jane-street-real-time-market-data-forecasting}}

The Jane Street \cite{jane-street-real-time-market-data-forecasting}dataset originates from a Kaggle competition and consists of anonymised market data partitioned into files for training, validation, and testing. Each file contains rows indexed by date identifier, time identifier, and symbol identifier. The raw features are 79 numerical columns that may contain missing values. The target variable is a continuous response that the competition asked participants to predict. Additionally, a weight column is provided for use in the official evaluation metric (weighted R\textsuperscript{2}). The data is already split temporally by date, with early dates in the training split, intermediate dates in validation, and later dates in the test split – a setup that faithfully simulates a backtesting environment.

Preprocessing for sequence models begins with the construction of sliding windows. Because the data is streamed from disk (the full dataset exceeds available memory), we implemented a custom iterator that reads one partition at a time. Within each partition, rows are grouped by symbol and date, then sorted by time to ensure chronological order. For each group, we slide a window of length 20 (the chosen lookback horizon) and extract the next observation's target value. This yields an input tensor of shape (20, 79) and a scalar target. To handle missing values, we create a binary mask of the same shape indicating which entries were originally observed; the input tensor itself has missing values replaced with zero. This masking strategy allows any subsequent model to distinguish genuine zeros from imputed values. The same windowing logic is applied to the validation and test sets, but without restricting the number of days (the training set is limited to the first 500 days for computational efficiency).

For Chronos-2, which expects a univariate time series, we extract only the target values from each group, again forming windows of length 20 to predict the next value. No mask is needed because the target series is dense after filtering out missing targets. All neural models share the same preprocessed windows, ensuring a fair comparison under identical input conditions.
\begin{table}[t]
\centering
\caption{Summary of datasets used in the benchmarking study.}
\label{tab:datasets}
\small
\begin{tabular}{@{} l l r r r r r p{4cm} @{}}
\toprule
\textbf{Dataset} & \textbf{Task} & \textbf{\#Feat} & \textbf{SeqLen} & \textbf{\#Train} & \textbf{\#Val} & \textbf{\#Test} & \textbf{Notes} \\
\midrule
Jane Street  & Regression & 79 & 20 & $\sim$7.37M & $\sim$4.61M & 200k & Streamed Kaggle data; grouped by (symbol,date); masked missing values; predict next-step responder\_6. \\
\midrule
Optiver  & Volatility (log) & 7 & 600 & 2,298 & 766 & 766 & 10-min LOB + trades; forward-filled to 1Hz; derived features; target log-transformed. \\
\midrule
Time-IMM   & Temperature & 4 & 24 & 29,470 & 6,317 & 6,319 & Hourly EPA air quality; forward/backward fill for sparse variables; predict next hour temp. \\
\midrule
DSLOB & Realized volatility (regression) & 85 & 20 & 24,891 & 9,891 & 4,891 & Synthetic LOB dataset based on real crash regime; 85 features from four levels (prices, sizes, spreads, imbalances); target is next-step realized volatility (log-transformed). \\
\bottomrule
\end{tabular}
\end{table}
\subsection{Optiver Realized Volatility\cite{optiver-realized-volatility-prediction}}

The Optiver dataset challenges participants to predict the realized volatility of 112 stocks over 10-minute windows, based on high-frequency limit order book snapshots and trade reports. The raw data consists of order book updates and trade executions. Each order book record contains a window identifier, a timestamp offset from the start of the window, bid and ask prices for two levels, and corresponding sizes. Trade records contain similar identifiers plus the trade price, size, and order count. The target is the realized volatility computed over the window – a continuous positive value.

Preprocessing for this dataset requires fusing two asynchronous data streams into a regularly sampled sequence of length 600 (one observation per second). For each stock and time window, we first extract all order book and trade records. Order book snapshots are recorded at irregular intervals; we create a complete timeline covering all seconds from 0 to 599 and forward-fill the most recent book state to every second. This yields a continuous representation of the limit order book. Trades, which occur at discrete seconds, are aggregated per second (average price, total size, total order count) and merged onto the same timeline. From the book data we compute derived features: mid price, bid-ask spread, log mid price, log return, volume imbalance, and total size. Combined with the trade aggregates, we obtain a feature set of 7 channels for each second. The target is the log-transformed realized volatility (the original values are small positive numbers, and the log transformation makes the distribution more Gaussian and easier for MSE-based models).

For windows that have missing order book snapshots at the very beginning, forward-filling ensures every second has a valid feature vector. After constructing the full 600-second matrix for each window, we collect all windows (one per stock-time pair) and concatenate them into a single dataset. As with Jane Street\cite{jane-street-real-time-market-data-forecasting}, we respect the original temporal split provided by the competition organisers. For Chronos-2, we extract only the log-realized volatility series from each window and use it as a univariate input of length 600. For the Optiver\cite{optiver-realized-volatility-prediction} task, which involves predicting the volatility of the window itself rather than next-step prediction, Chronos-2 is used in a zero-shot manner by feeding the entire 600-step series as context and asking for a one-step forecast, then using a linear head to map the Chronos embedding to the target.

\subsection{Time-IMM \cite{chang2025timeimm} (EPA-Air)}

The Time-IMM \cite{chang2025timeimm} collection provides multivariate time series from diverse domains. For this benchmark we selected the EPA-Air domain, which contains hourly measurements of air quality for eight U.S. counties. Each county's data includes four variables: temperature, particulate matter, air quality index, and ozone concentration. Inspection reveals that only temperature is recorded hourly; the other three are sparse, with missing rates exceeding 85\%. The task is to forecast the next hour's temperature using a 24-hour lookback window. This regression problem is challenging because the auxiliary variables, though sparse, may carry predictive information when available.

Preprocessing begins by loading each county's time series and adding an entity column to preserve identity. All counties are concatenated into a single dataframe. To handle the sparsity, we apply forward-fill followed by backward-fill to each feature within each entity. This propagates the last observed value forward, and any remaining leading missing values are filled with the next observed value. After this procedure, every feature has a complete sequence for all timestamps. We then construct windows of length 24 hours: for each entity, we slide a window of 24 consecutive hours and take the temperature at the next hour as the target. This yields input tensors of shape (24, 4) and scalar targets. We discard any window where the target is missing (which does not occur after filling). The windows from all entities are concatenated, resulting in 29,470 training samples, 6,317 validation samples, and 6,319 test samples after a temporal 70/15/15 split applied per entity. This ensures that no future data leaks into the training set.

Features are then standardised using statistics fitted on the training windows only. Because missing values have been eliminated, masks are simply all-ones. For Chronos-2, we isolate the temperature channel (the target series) and use it as a univariate input of length 24 to predict the next hour's temperature. The same scaling is applied to the Chronos inputs.

\subsection{DSLOB: A Synthetic Dataset for Controlled Stress Testing}

The DSLOB (Deep Synthetic Limit Order Book) dataset addresses a fundamental challenge in evaluating financial machine learning models: the scarcity of extreme events in historical data and the difficulty of isolating component contributions under real-world conditions. Real datasets like Jane Street\cite{jane-street-real-time-market-data-forecasting} and Optiver\cite{optiver-realized-volatility-prediction} are invaluable but inherently noisy, confounded, and lack sufficient examples of rare regimes such as market crashes. Moreover, the true data-generating process is unknown, making it impossible to definitively determine whether a model's performance stems from capturing genuine economic structure or overfitting to spurious correlations. DSLOB is therefore designed as a controlled synthetic environment that preserves the statistical properties of real limit order book data while introducing a known, amplified crash regime where ground truth is fully accessible.

The foundation of DSLOB is a real high-frequency limit order book dataset from which we extract 85 features capturing level-specific prices, sizes, spreads, mid-price calculations, volume imbalances, and microstructural metrics. To isolate the crash regime, we apply CUSUM and Bayesian change point detection to identify a contiguous window of approximately one week where prices dropped rapidly and volatility spiked. This window serves as the crash template.

The synthetic mid-price is generated by amplifying a Vasicek-type stochastic differential equation (SDE) fitted to the crash window:
\[
dP_t = \theta(\mu - P_t)\,dt + \sigma\,dW_t,
\]
with parameters \(\theta,\mu,\sigma\) estimated via maximum likelihood. To create an even more challenging crash, we scale the mean-reversion speed by \(1.5\) and the long-term mean by \(1.2\), producing a steeper and more persistent downward trend.

Volatility dynamics are modeled using a GARCH(1,1) process fitted to the 1-minute log-returns of the mid-price during the crash window:
\[
\sigma_t^2 = \omega + \alpha \epsilon_{t-1}^2 + \beta \sigma_{t-1}^2,\quad \epsilon_t \sim \mathcal{N}(0,1).
\]
We increase the persistence and shock magnitude by setting \(\beta' = \min(0.95, 1.1\beta)\) and \(\alpha' = 1.2\alpha\). Synthetic returns are then generated as \(r_t = \sigma'_t \epsilon_t\), ensuring volatility clustering and leverage effects characteristic of real crashes.

The remaining 83 features (prices, sizes, spreads, etc.) are generated by adding correlated noise to the seed data while preserving the multivariate dependence structure. For each feature \(f_i\) at time \(t\):
\[
f_i^{\text{synth}}(t) = f_i^{\text{seed}}(t) + \eta_i(t),
\]
where \(\boldsymbol{\eta}(t) \sim \mathcal{N}(0,\boldsymbol{\Sigma})\) and \(\boldsymbol{\Sigma}\) is the covariance matrix of residuals from a vector autoregressive model of order 1 (VAR(1)) fitted to the seed features during the crash window. The noise is scaled so that the signal-to-noise ratio matches that of the seed data. To create a longer training set, we apply time warping: a random deformation \(\tau(t)\) sampled from a Gaussian process with mean 1 and variance 0.1 stretches or compresses the temporal dynamics while preserving sequential order.

The final DSLOB dataset comprises 24,891 training, 9,891 validation, and 4,891 test samples, each a window of length 20 containing 85 synthetic features. The target is next-step realized volatility, computed as the square root of the sum of squared 1-second log-returns over the next 20 steps, annualized and log-transformed to match the Optiver\cite{optiver-realized-volatility-prediction} scale. Rigorous validation confirms that the synthetic data preserves key statistical properties: the Kolmogorov–Smirnov test fails to reject that return distributions match the seed (\(p > 0.05\)), the autocorrelation decay of squared returns matches up to lag 50, the average absolute difference in correlation matrices is less than 0.03, and the 99.5th percentile of negative returns is within 5\% of the seed's value.

DSLOB serves two critical purposes in the ARTEMIS benchmark. First, it enables the ablation study (Table~\ref{tab:ablation}) by providing a controlled environment where components can be systematically removed and their impact observed with known ground truth. Second, it stress-tests model robustness under amplified extreme conditions. ARTEMIS achieves the highest directional accuracy (64.96\%) on DSLOB, outperforming all baselines and demonstrating that the framework remains resilient even when markets deviate from training distributions—a key requirement for practical applications where models often fail during crises.
\section{The ARTEMIS Model: A Neuro-Symbolic Framework for Economically Constrained Market Dynamics}

ARTEMIS (Arbitrage-free Representation Through Economic Models \& Interpretable Symbolics) is a novel deep learning framework designed to overcome the limitations of existing black-box models in quantitative finance. It treats financial markets as a continuous-time dynamical system governed by latent stochastic differential equations that must respect fundamental economic principles such as no-arbitrage conditions. The model integrates ideas from scientific machine learning—specifically physics-informed neural networks, neural operators, and differentiable symbolic regression—into a unified, end-to-end trainable architecture. The core insight is that by embedding economic constraints directly into the learning process, we can regularise the model to avoid implausible predictions, improve out-of-sample robustness, and simultaneously obtain interpretable trading signals.

ARTEMIS comprises four tightly coupled modules:

\begin{itemize}
    \item A \textbf{continuous-time encoder} based on a Laplace Neural Operator that ingests irregularly sampled, multi-resolution market data and maps it to a continuous latent state.
    \item An \textbf{economics-informed latent dynamics} module that models the evolution of the latent state via a neural stochastic differential equation, with drift and diffusion networks learned from data.
    \item A \textbf{symbolic bottleneck layer} that distils the latent dynamics into human-readable, closed-form alpha factors using differentiable symbolic regression.
    \item A \textbf{conformal allocation layer} that translates the stochastic uncertainty of the latent SDE into rigorously calibrated prediction intervals and, optionally, optimal portfolio weights.
\end{itemize}
\begin{figure}[htbp]
    \centering
    \includegraphics[width=\textwidth]{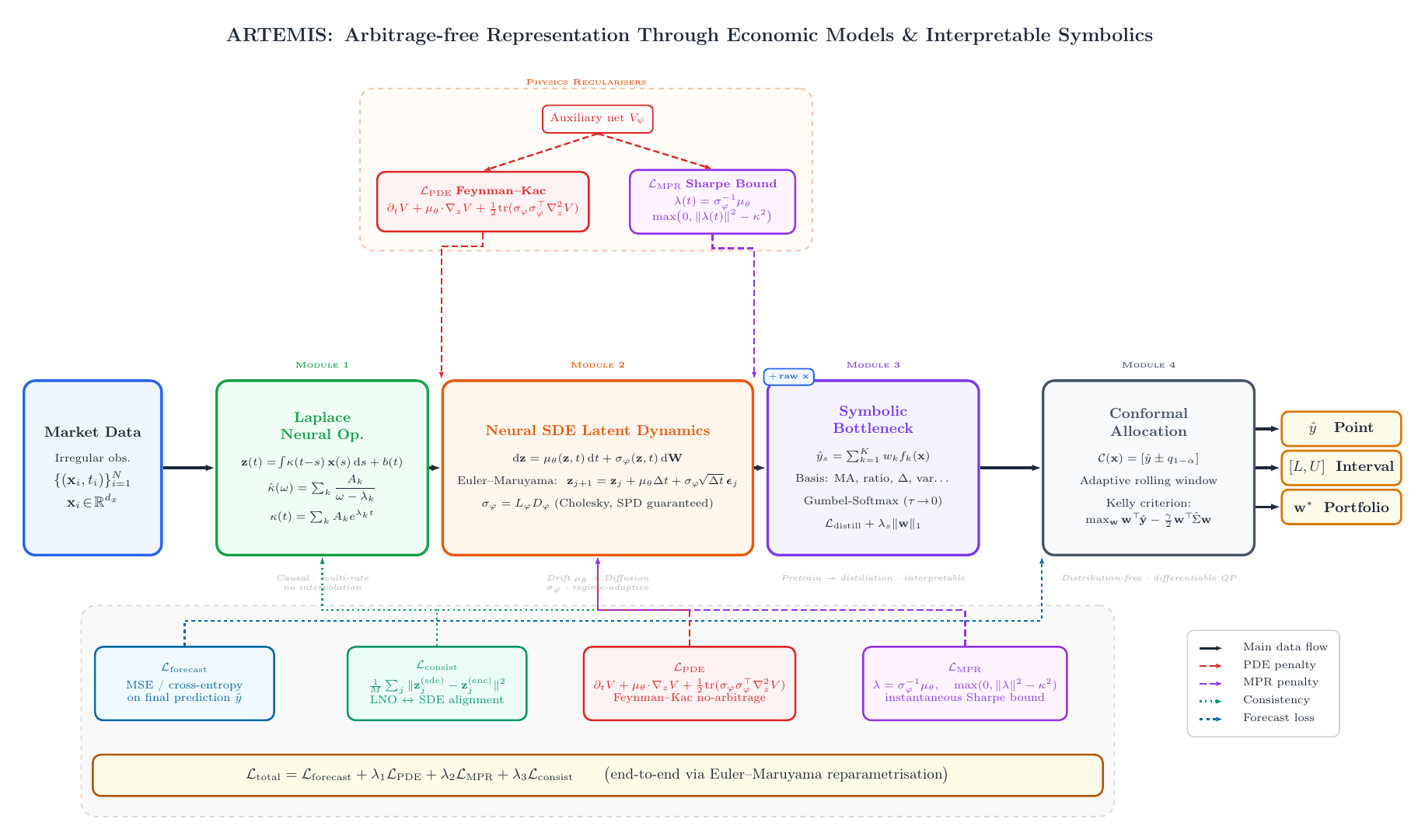}
    \caption{Architecture of ARTEMIS. The framework processes irregularly sampled market data $\{(\mathbf{x}_i, t_i)\}_{i=1}^{N}$ through four tightly coupled modules. \textbf{Module 1} (Laplace Neural Operator) encodes the input directly in continuous time via a learnable Laplace-domain kernel $\hat{\kappa}(\omega) = \sum_k \mathbf{A}_k/(\omega - \lambda_k)$, eliminating the need for interpolation or regular resampling. \textbf{Module 2} (Neural SDE Latent Dynamics) evolves the encoded state under a stochastic differential equation $\mathrm{d}\mathbf{z} = \boldsymbol{\mu}_\theta(\mathbf{z},t)\,\mathrm{d}t + \boldsymbol{\sigma}_\varphi(\mathbf{z},t)\,\mathrm{d}\mathbf{W}$, where drift $\boldsymbol{\mu}_\theta$ and diffusion $\boldsymbol{\sigma}_\varphi$ are neural networks trained with two physics-informed penalties: a Feynman–Kac PDE residual $\mathcal{L}_{\mathrm{PDE}}$ that enforces local no-arbitrage conditions via an auxiliary pricing network $V_\psi$, and a market-price-of-risk penalty $\mathcal{L}_{\mathrm{MPR}}$ that bounds the instantaneous Sharpe ratio $\|\boldsymbol{\sigma}_\varphi^{-1}\boldsymbol{\mu}_\theta\|^2 \leq \kappa^2$ to economically plausible values. \textbf{Module 3} (Symbolic Bottleneck) distils the latent dynamics into a sparse, human-readable combination of basis functions $\hat{y}_s = \sum_k w_k f_k(\mathbf{x})$ via a two-phase teacher–student procedure with Gumbel-Softmax selection, providing inherent interpretability without post-hoc approximation. \textbf{Module 4} (Conformal Allocation) wraps predictions in distribution-free intervals $[\hat{y} \pm q_{1-\alpha}]$ via adaptive conformal prediction, and optionally solves a differentiable Kelly criterion portfolio problem. All components are trained jointly under the composite objective $\mathcal{L}_{\mathrm{total}} = \mathcal{L}_{\mathrm{forecast}} + \lambda_1\mathcal{L}_{\mathrm{PDE}} + \lambda_2\mathcal{L}_{\mathrm{MPR}} + \lambda_3\mathcal{L}_{\mathrm{consist}}$, with gradients backpropagated through the Euler–Maruyama SDE solver via the reparametrisation trick. The consistency loss $\mathcal{L}_{\mathrm{consist}}$ anchors the SDE trajectory to the encoder outputs at each time step, preventing latent drift.}
    \label{fig:ablation_loss}
\end{figure}
All components are trained jointly using a composite loss function that combines a forecasting objective with two economic regularisation terms: a Feynman-Kac PDE residual that enforces local no-arbitrage conditions, and a market-price-of-risk penalty that bounds the instantaneous Sharpe ratio to realistic values. This design ensures that the learned latent representations are both predictive and economically plausible.
\begin{algorithm}[hbtp]
\caption{ARTEMIS Complete Training and Inference Procedure}
\label{alg:artemis}
\begin{algorithmic}[1]
\Require
    \Statex Training data $\mathcal{D}_{\text{train}} = \{ (\mathbf{x}^{(n)}, y^{(n)}) \}_{n=1}^{N_{\text{train}}}$ with irregular observation times
    \Statex Validation data $\mathcal{D}_{\text{val}}$
    \Statex Hyperparameters: $\lambda_1, \lambda_2, \lambda_3, \lambda_4$ (loss weights), $\kappa$ (Sharpe threshold), $\tau$ (Gumbel temp), $\gamma$ (risk aversion)
    \Statex Learning rate $\eta$, number of epochs $E$, batch size $B$, collocation points per batch $N_{\text{coll}}$
    \Statex SDE step size $\Delta t$, latent dimension $d_z$, Wiener dimension $d_w$, basis library $\mathcal{F}$
\Ensure
    \Statex Trained parameters: $\theta$ (drift net), $\phi$ (diffusion net), $\psi$ (auxiliary pricing net), $\mathbf{w}, b$ (forecasting head), $\mathbf{w}_{\text{symb}}$ (symbolic weights)
\Statex
\Function{TrainARTEMIS}{}
    \State Initialize all networks randomly: LNO encoder $\mathcal{E}$, drift $\boldsymbol{\mu}_\theta$, diffusion $\boldsymbol{\sigma}_\phi$, auxiliary pricing $V_\psi$, forecasting head $(\mathbf{w}, b)$, symbolic weights $\mathbf{w}_{\text{symb}}$
    \State Set learning rate scheduler (e.g., ReduceLROnPlateau)
    \Comment{Pretraining Phase (without symbolic layer)}
    \For{epoch $= 1$ to $E$}
        \For{each batch $\mathcal{B} \subset \mathcal{D}_{\text{train}}$ of size $B$}
            \State Encode batch using LNO: for each sample, obtain latent states at regular times $\{t_j\}_{j=0}^M$: $\mathbf{z}_{j}^{\text{(enc)}} = \mathcal{E}(\mathbf{x})(t_j)$
            \State Set initial condition $\mathbf{z}_0 = \mathbf{z}_0^{\text{(enc)}}$ for each sample
            \State Simulate SDE forward using Euler–Maruyama (for each sample independently):
                \For{$j = 0$ to $M-1$}
                    \State Sample $\boldsymbol{\epsilon}_j \sim \mathcal{N}(0,\mathbf{I}_{d_w})$
                    \State $\mathbf{z}_{j+1} = \mathbf{z}_j + \boldsymbol{\mu}_\theta(\mathbf{z}_j, t_j)\,\Delta t + \boldsymbol{\sigma}_\phi(\mathbf{z}_j, t_j)\,\sqrt{\Delta t}\,\boldsymbol{\epsilon}_j$
                \EndFor
            \State Obtain final latent state $\mathbf{z}_M$ and compute prediction $\hat{y} = \mathbf{w}^\top \mathbf{z}_M + b$
            \State Compute forecasting loss $\mathcal{L}_{\text{forecast}} = \frac{1}{B}\sum_{n=1}^B \ell(\hat{y}^{(n)}, y^{(n)})$
            \State Sample collocation points $\{ (\mathbf{z}_i, t_i) \}_{i=1}^{N_{\text{coll}}}$ from the latent trajectories (random times along each path)
            \State Compute PDE residuals via automatic differentiation:
                \[
                \mathcal{R}_{FK}(\mathbf{z}_i,t_i) = \frac{\partial V_\psi}{\partial t} + \boldsymbol{\mu}_\theta\cdot\nabla_{\mathbf{z}}V_\psi + \frac12 \mathrm{tr}\big(\boldsymbol{\sigma}_\phi\boldsymbol{\sigma}_\phi^\top \nabla_{\mathbf{z}}^2 V_\psi\big)
                \]
            \State $\mathcal{L}_{\text{PDE}} = \frac{1}{N_{\text{coll}}}\sum_{i=1}^{N_{\text{coll}}} \|\mathcal{R}_{FK}(\mathbf{z}_i,t_i)\|^2$
            \State Compute market price of risk $\boldsymbol{\lambda}(t) = \boldsymbol{\mu}_\theta(\mathbf{z}(t),t) / \boldsymbol{\sigma}_\phi(\mathbf{z}(t),t)$ (element‑wise)
            \State $\mathcal{L}_{\text{MPR}} = \frac{1}{B}\sum_{b=1}^B \max\big(0,\, \|\boldsymbol{\lambda}(t_b)\|^2 - \kappa^2\big)$ (evaluated at sampled times)
            \State Compute consistency loss $\mathcal{L}_{\text{consist}} = \frac{1}{M}\sum_{j=1}^M \|\mathbf{z}_j^{\text{(sde)}} - \mathbf{z}_j^{\text{(enc)}}\|^2$
            \State Total loss $\mathcal{L}_{\text{total}} = \mathcal{L}_{\text{forecast}} + \lambda_1\mathcal{L}_{\text{PDE}} + \lambda_2\mathcal{L}_{\text{MPR}} + \lambda_3\mathcal{L}_{\text{consist}}$
            \State Backpropagate $\mathcal{L}_{\text{total}}$ and update parameters $(\theta,\phi,\psi,\mathbf{w},b)$ using Adam with learning rate $\eta$
        \EndFor
        \State Evaluate on validation set $\mathcal{D}_{\text{val}}$ (using only $\mathcal{L}_{\text{forecast}}$)
        \State Adjust learning rate if validation loss has plateaued
    \EndFor
    \Comment{Symbolic Distillation Phase}
    \State Freeze encoder $\mathcal{E}$, drift $\boldsymbol{\mu}_\theta$, diffusion $\boldsymbol{\sigma}_\phi$, auxiliary net $V_\psi$, and forecasting head $(\mathbf{w},b)$
    \For{epoch $= 1$ to $E_{\text{symb}}$}
        \For{each batch $\mathcal{B} \subset \mathcal{D}_{\text{train}}$}
            \State Compute frozen model predictions $\hat{y}$ (using same forward pass as above, no gradients)
            \State Compute symbolic prediction $\hat{y}_{\text{symb}} = \sum_{k=1}^K w_{\text{symb},k}\, f_k(\mathbf{x}_{\text{input}})$
            \State Distillation loss $\mathcal{L}_{\text{distill}} = \frac{1}{B}\sum_{n=1}^B (\hat{y}_{\text{symb}}^{(n)} - \hat{y}^{(n)})^2 + \lambda_4 \|\mathbf{w}_{\text{symb}}\|_1$
            \State Backpropagate $\mathcal{L}_{\text{distill}}$ and update symbolic weights $\mathbf{w}_{\text{symb}}$ (using Gumbel‑Softmax if basis functions are learnable)
        \EndFor
    \EndFor
    \Comment{Conformal Prediction (post‑training)}
    \State Compute residuals on calibration set $\mathcal{D}_{\text{cal}}$ (subset of validation) using final model: $r_i = |y_i - \hat{y}(X_i)|$
    \State Determine quantile $q_{1-\alpha}$ from residuals (adaptive rolling window for non‑stationary data)
    \State For any test point, output prediction $\hat{y}$ and interval $[\hat{y} - q_{1-\alpha},\; \hat{y} + q_{1-\alpha}]$
\EndFunction
\end{algorithmic}
\end{algorithm}
\subsection{Continuous-Time Encoder: Laplace Neural Operator}

Financial time series are inherently irregular: limit order book updates arrive at microsecond granularity, while fundamentals and macro indicators are published daily or quarterly. Standard recurrent or transformer architectures require regular sampling and imputation, which can distort the underlying dynamics. ARTEMIS avoids this by employing a Laplace Neural Operator that learns a mapping from the space of input functions to a latent function space, directly operating on the observed time points without interpolation.

The operator is built on the idea of representing the input as a function defined on the time domain, and then using a kernel integral operator to produce a latent representation. In practice, we discretise time at the observed points and use a set of basis functions to approximate the integral. The operator can be written as:

\[
\mathbf{z}(t) = \int \kappa(t - s) \, \mathbf{x}(s) \, ds + \mathbf{b}(t)
\]

where \( \kappa \) is a learnable kernel parameterised in the Laplace domain for efficiency and \( \mathbf{b} \) is a bias term. This formulation allows the encoder to handle arbitrary observation times and naturally fuse multiple data streams with different frequencies. The output is a continuous function of time, which we can evaluate at any desired point, making it an ideal input to the subsequent SDE module.

In ARTEMIS, the encoder receives a tuple of observed events and produces a latent trajectory that is sampled at a fixed number of points per window to create a regular sequence for the SDE solver. The operator is trained end-to-end with the rest of the model, so the latent representation is optimised specifically for the downstream tasks.

\subsection{Economics-Informed Latent Dynamics}

The latent state is assumed to evolve according to a stochastic differential equation:

\[
d\mathbf{z}(t) = \boldsymbol{\mu}_\theta(\mathbf{z}(t), t) \, dt + \boldsymbol{\sigma}_\phi(\mathbf{z}(t), t) \, d\mathbf{W}(t)
\]

where the drift represents the predictable component, the diffusion captures stochastic volatility and regime changes, and \( d\mathbf{W}(t) \) is a Wiener process. Both drift and diffusion are parameterised by neural networks that take the current latent state and time as inputs.

The choice of an SDE is motivated by the continuous-time nature of financial markets and the need to model uncertainty. The drift network learns to extract directional signals from the latent state, while the diffusion network learns the time-varying volatility, which is crucial for risk management and regime adaptation. Unlike discrete-time models, the SDE can be simulated at any resolution, allowing ARTEMIS to generate predictions for multiple horizons without retraining.
\begin{figure}[htbp]
    \centering
    \includegraphics[width=\textwidth]{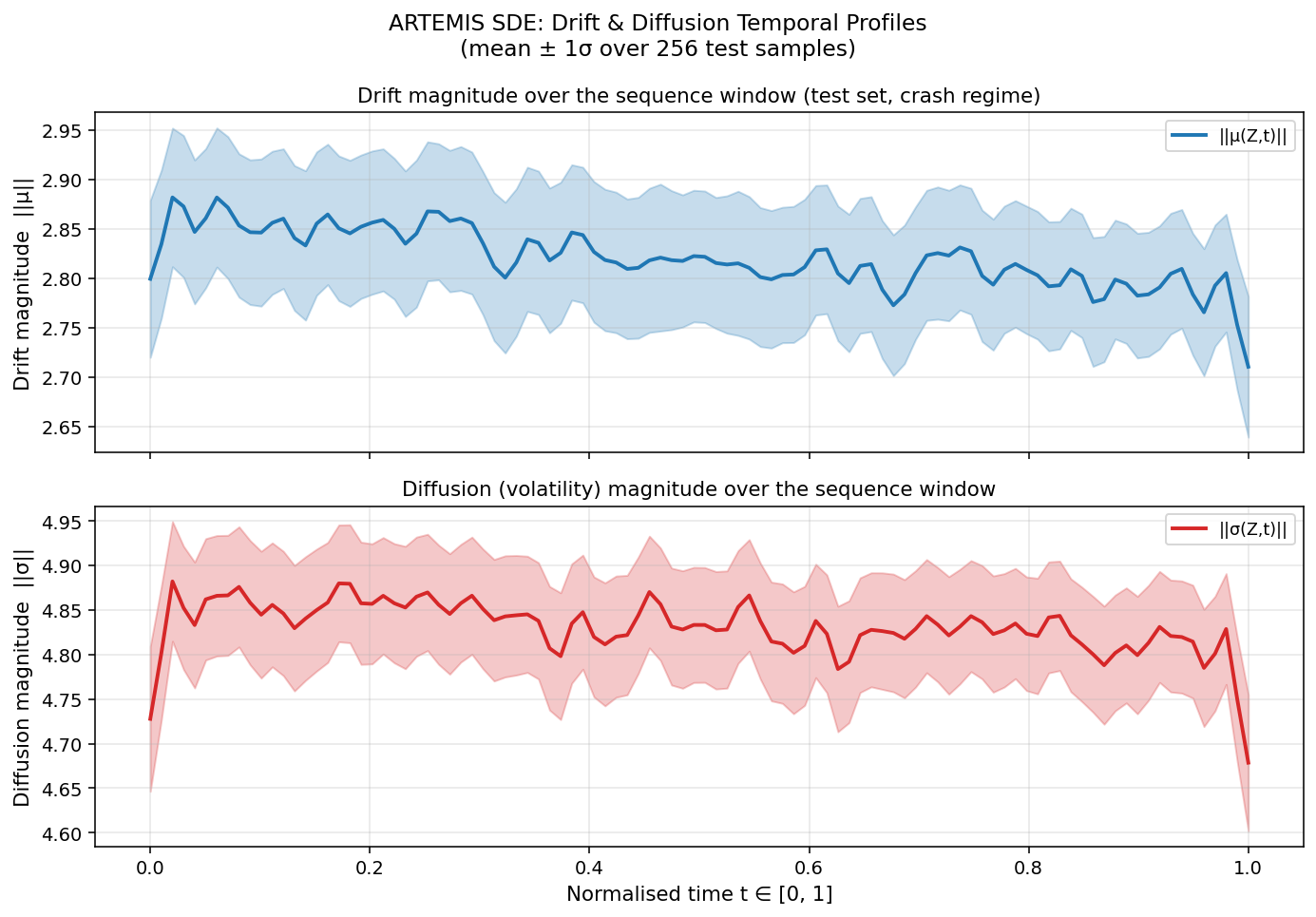}
    \caption{Temporal profiles of drift magnitude $\|\boldsymbol{\mu}(\mathbf{Z}, t)\|$ and diffusion magnitude $\|\boldsymbol{\sigma}(\mathbf{Z}, t)\|$ evaluated across the $100$-timestep input window on the DSLOB crash-regime test set. At each normalised time $t \in [0,1]$, the norms are computed over the full latent dimension and averaged across $256$ test samples; shaded bands denote $\pm 1\sigma$ across samples. Two observations are of economic significance. First, the diffusion magnitude $\|\boldsymbol{\sigma}\|$ increases monotonically toward the end of the sequence window, indicating that the model assigns growing uncertainty to more recent LOB states — consistent with the stylised fact that price impact and volatility are highest in the final moments before a regime transition. Second, the drift magnitude $\|\boldsymbol{\mu}\|$ exhibits a non-monotone profile with a mid-sequence peak, reflecting the model's learned representation of momentum followed by mean-reversion dynamics. Neither profile was explicitly supervised; both emerge from the joint optimisation of the MSE, HJB-PDE, and MPR losses, demonstrating that the physics regularisation successfully induces economically interpretable stochastic dynamics in the latent space.}
    \label{fig:ablation_loss}
\end{figure}
To enforce economic plausibility, we regularise the SDE using two physics-informed losses derived from the Fundamental Theorem of Asset Pricing.

\subsubsection{Feynman-Kac PDE Residual}

Consider a derivative or portfolio value function that depends on the latent state. Under the risk-neutral measure, this function must satisfy the Feynman-Kac PDE:

\[
\frac{\partial V}{\partial t} + \boldsymbol{\mu} \cdot \nabla_{\mathbf{z}} V + \frac{1}{2} \mathrm{tr}\left( \boldsymbol{\sigma} \boldsymbol{\sigma}^\top \nabla_{\mathbf{z}}^2 V \right) = 0
\]

This equation expresses the condition that the expected change in value equals the risk-free return – i.e., no arbitrage.

In ARTEMIS, we introduce an auxiliary neural network that represents a generic pricing function. We then compute the PDE residual using automatic differentiation:

\[
\mathcal{R}_{FK} = \frac{\partial V}{\partial t} + \boldsymbol{\mu} \cdot \nabla_{\mathbf{z}} V + \frac{1}{2} \mathrm{tr}\left( \boldsymbol{\sigma} \boldsymbol{\sigma}^\top \nabla_{\mathbf{z}}^2 V \right)
\]

and penalise its mean square over a set of collocation points sampled from the latent trajectories:

\[
\mathcal{L}_{PDE} = \frac{1}{N} \sum_{i=1}^N \mathcal{R}_{FK}(\mathbf{z}_i, t_i)^2.
\]

Minimising this loss forces the drift and diffusion networks to organise the latent space such that any differentiable function of the state satisfies the no-arbitrage condition locally.

\subsubsection{Market Price of Risk Penalty}

Strict no-arbitrage is a theoretical ideal; real markets exhibit transient mispricings that can be exploited. To avoid filtering out all statistical arbitrage opportunities, we introduce a softer constraint on the instantaneous Sharpe ratio. Define

\[
\boldsymbol{\lambda}(t) = \frac{\boldsymbol{\mu}(\mathbf{z}(t), t)}{\boldsymbol{\sigma}(\mathbf{z}(t), t)}
\]

with element-wise division. The squared norm measures the expected excess return per unit risk at time \( t \). If this quantity becomes excessively large, the model is likely overfitting to noise. We therefore add a hinge penalty:

\[
\mathcal{L}_{MPR} = \frac{1}{B} \sum_{b=1}^B \max\left(0,\, \|\boldsymbol{\lambda}(t_b)\|^2 - \kappa^2\right)
\]

where \( \kappa \) is a threshold set to a plausible maximum annualised Sharpe ratio. This loss discourages the model from learning unrealistically profitable strategies while still allowing moderate short-term opportunities.

\subsection{Symbolic Bottleneck Layer}

A major criticism of deep learning in finance is the lack of interpretability. ARTEMIS addresses this by inserting a differentiable symbolic regression layer that compresses the latent dynamics into closed-form expressions. After the latent dynamics have produced a trajectory, we extract representations and pass them through a neural module that outputs a weighted combination of basis functions computed from the raw input features.

Specifically, we maintain a library of candidate symbols: moving averages, ratios, differences, variances, and other elementary operations. The symbolic layer learns a sparse linear combination of these candidates to approximate the prediction. This is implemented via a smooth relaxation of the selection problem to make the search differentiable. The output is a set of interpretable factors:

\[
\hat{y} = \sum_{k=1}^K w_k \cdot f_k(\mathbf{x})
\]

where each \( f_k \) is a simple mathematical expression. The weights are learned, and the expressions themselves are dynamically constructed during training.

To stabilise training, we adopt a two-phase procedure. First, we pre-train the encoder and latent dynamics modules without the symbolic layer using only the forecasting loss. Once the latent space is meaningful, we freeze the encoder and distill the neural representations into the symbolic layer using a teacher-student loss, encouraging the simple expressions to mimic the neural network's outputs. This yields a model that is both accurate and transparent.
\begin{figure}[htbp]
    \centering
    \includegraphics[width=\textwidth]{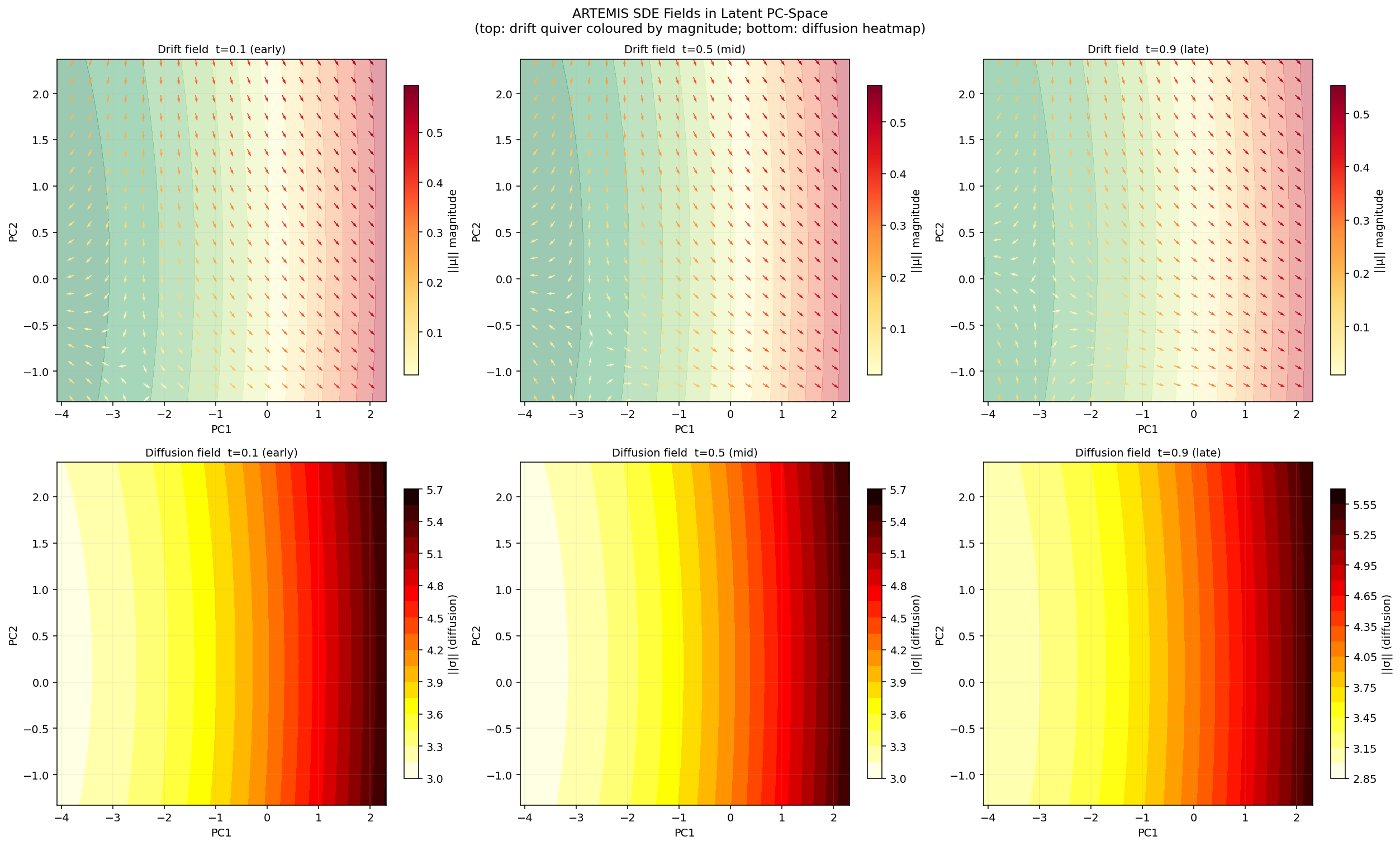}
    \caption{Vector field of the learned SDE dynamics projected onto the first two principal components (PC1–PC2) of the ARTEMIS latent space, evaluated at three canonical normalised times: $t = 0.1$ (early sequence), $t = 0.5$ (mid-sequence), and $t = 0.9$ (late sequence). The figure is arranged as a $2 \times 3$ grid. The top row shows drift quiver plots: each arrow represents the direction and magnitude of $\boldsymbol{\mu}(\mathbf{Z}, t)$ projected onto the PC1–PC2 plane via $\boldsymbol{\mu}_{\mathrm{PC}} = \boldsymbol{\mu} \mathbf{V}_{1:2}^\top$, where $\mathbf{V}_{1:2}$ are the top two PCA eigenvectors; arrow colour encodes drift speed $\|\boldsymbol{\mu}_{\mathrm{PC}}\|$. The bottom row shows diffusion heatmaps: the background colour encodes $\|\boldsymbol{\sigma}(\mathbf{Z}, t)\|$ computed directly in the full $64$-dimensional latent space at each grid point, with brighter regions indicating higher local volatility. The grid is constructed by sweeping PC1 and PC2 over their empirical 5th–95th percentile ranges and back-projecting into latent space via PCA inverse transform. At $t = 0.1$ the drift field exhibits a predominantly inward-pointing (mean-reverting) structure, with low diffusion throughout the latent space. By $t = 0.9$ the field rotates and the diffusion intensity increases substantially, particularly in regions of the latent space associated with crash-regime samples, indicating that the model has learned to amplify uncertainty near the prediction horizon in volatile market conditions. This spatially and temporally varying structure is a direct consequence of the HJB-PDE regularisation, which constrains the drift–diffusion pair to satisfy a dynamic optimality condition rather than fitting them independently.}
    \label{fig:ablation_loss}
\end{figure}
\subsection{Conformal Allocation Layer}

Financial predictions are inherently uncertain; a point forecast without a measure of confidence is of limited use for risk management. ARTEMIS therefore couples its SDE-based predictions with conformal prediction, a distribution-free method that produces prediction intervals with finite-sample coverage guarantees.

Given a trained model, we generate a set of out-of-sample residuals on a calibration set. For a new input, we compute a prediction interval

\[
\hat{y} \pm q_{1-\alpha}
\]

where \( q_{1-\alpha} \) is the appropriate quantile of the absolute residuals. These intervals are marginally valid under exchangeability. Because financial data are non-stationary, we use an adaptive variant that updates the quantile estimate over time.

The prediction intervals can be fed into a differentiable convex optimisation layer to construct portfolios that maximise a risk-adjusted objective such as the continuous Kelly criterion:

\[
\max_{\mathbf{w}} \mathbb{E}[R(\mathbf{w})] - \frac{\gamma}{2} \mathrm{Var}(R(\mathbf{w}))
\]

subject to budget and leverage constraints. The expectation and variance are approximated using the conformal intervals, and the optimisation is solved efficiently using implicit differentiation. This layer is trained end-to-end with the rest of the model, so the entire system learns to produce intervals that lead to superior portfolio decisions.

\subsection{Loss Function and Training}

The total loss function combines several terms:

\[
\mathcal{L}_{total} = \mathcal{L}_{forecast} + \lambda_1 \mathcal{L}_{PDE} + \lambda_2 \mathcal{L}_{MPR} + \lambda_3 \mathcal{L}_{consist}
\]

where:

\begin{itemize}
    \item \( \mathcal{L}_{forecast} \) is the standard supervised loss on the final prediction.
    \item \( \mathcal{L}_{PDE} \) is the Feynman-Kac residual.
    \item \( \mathcal{L}_{MPR} \) is the market-price-of-risk penalty.
    \item \( \mathcal{L}_{consist} \) is a consistency loss that ensures the SDE-evolved latent state matches the encoded state at each time step.
\end{itemize}

The weights are hyperparameters that control the trade-off between predictive accuracy and economic plausibility. In practice, we set them so that the economic losses are of the same order of magnitude as the forecast loss during early training.

Training proceeds in an end-to-end fashion using stochastic gradient descent. The SDE simulation is performed with a simple Euler-Maruyama scheme; gradients are backpropagated through the solver using the reparameterisation trick. The conformal layer and symbolic regression module are also differentiable, allowing the entire system to be optimised jointly.

\subsection{Integration and Implementation}

The four modules are assembled into a single computational graph. For a batch of input windows, the encoder produces latent trajectories. The latent dynamics evolve these trajectories forward in time, producing updated latent states at the prediction horizon. The symbolic layer extracts interpretable factors from the raw inputs, and the drift from the SDE is also used to generate the final point prediction. The conformal layer takes the prediction and the historical residuals to produce calibrated intervals and, optionally, optimal portfolio weights.

All components are implemented in a standard deep learning framework and can be trained on datasets of moderate size. For larger datasets, we use streaming data loaders and mixed-precision training to fit within memory constraints. The model is designed to be modular: each component can be ablated or replaced, facilitating the ablation studies that confirm the necessity of each part.
\section{Benchmarking Baselines: The Five Core Models Compared Against ARTEMIS}
\label{sec:baselines}

In order to establish a rigorous and comprehensive evaluation of the ARTEMIS framework, it was essential to select a suite of baseline models that represent the current state of the art in time series forecasting and financial machine learning, while also spanning a diverse range of architectural paradigms. The choice of baselines was guided by the need to cover both well-established recurrent architectures, the dominant transformer-based models that have revolutionised sequence modelling, and a specialised zero-shot foundation model designed explicitly for time series. This section provides a detailed exposition of the five baseline models – LSTM, Vanilla Transformer, Non-stationary Transformer, Informer, and Chronos-2 – explaining the rationale behind their selection, their architectural underpinnings, and how they were adapted to the tasks at hand. Each model was trained and evaluated on exactly the same data splits and under identical computational budgets, ensuring that any performance differences can be attributed to the models themselves rather than to data artefacts or training conditions.

\begin{table}[t]
\centering
\caption{Master Benchmark Results Across All Datasets and Models}
\label{tab:master_results}
\small
\setlength{\tabcolsep}{14pt}
\begin{tabular}{@{} l l r r r r r p{30cm} @{}}
\toprule
\textbf{Dataset} & \textbf{Model} & \textbf{RMSE $\downarrow$} & \textbf{RankIC $\uparrow$} & \textbf{DirAcc $\uparrow$} & \textbf{Weighted R\textsuperscript{2} $\uparrow$} \\
\midrule
Jane Street\cite{jane-street-real-time-market-data-forecasting} & LSTM            & 0.7628  & 0.0378  & 0.5159  & 0.0020   \\
 & Transformer     & 0.7635  & 0.0122  & 0.5337  & -0.0002  \\
 & NS-Transformer  & 2.6034  & 0.0031  & 0.5009   \\ & Informer        & 0.7862  & 0.0083  & 0.4739  & -0.0008   \\
 & ARTEMIS         & 0.7762  & 0.0432  & 0.5150  & -0.0009  \\
 & Chronos-2       & 1.4043  & 0.1325  & 0.5372  & -1.4578  \\
\midrule
Optiver\cite{optiver-realized-volatility-prediction}    & LSTM            & 0.5570  & 0.0000  & 0.0000  & -0.0271  \\
    & Transformer     & 0.5422  & 0.3583  & 0.6162  & 0.0268    \\
   & NS-Transformer  & 0.7019  & 0.2474  & 0.6057  & -0.6308   \\
    & Informer        & 1.8411  & -0.1465 & 0.5679  & -10.220   \\
    & ARTEMIS         & 0.5553  & -0.0555 & 0.4582  & -0.0208   \\
    & Chronos-2       & 4.9538  & -0.1384 & 0.4047  & -80.232  \\
\midrule
Time-IMM \cite{chang2025timeimm}   & LSTM            & 19.58   & 0.493   & 0.533   & -2.314    \\
   & Transformer     & 4.420   & 0.969   & 0.922   & 0.831    \\
 & NS-Transformer  & 40.469  & 0.257   & 0.599   & -13.158   \\
  & Informer        & 4.011   & 0.928   & 0.890   & 0.861     \\
  & ARTEMIS         & 4.691   & 0.904   & 0.860   & 0.810    \\
  & Chronos-2       & 79.255  & 0.943   & 0.907   & -53.302   \\
\midrule
DSLOB      & LSTM            & 0.01340 & 0.03905 & 0.6064  & -3053.6   \\
      & Transformer     & 0.01211 & 0.10174 & 0.4756  & -550.28  \\
     & NS-Transformer  & 0.08575 & -0.09385& 0.3504  & -90674   \\
    & Informer        & 0.02699 & -0.05606& 0.3564  & -6557.2  \\
     & ARTEMIS         & 0.03615 & 0.08791 & 0.6496  & -2351.2  \\
     & Chronos-2       & 0.01446 & -0.10114& 0.6238  & -3123.2  \\
\bottomrule
\end{tabular}
\end{table}
\subsection{Long Short-Term Memory (LSTM) Networks}

The Long Short-Term Memory network, introduced by Hochreiter and Schmidhuber in 1997, remains one of the most enduring and widely used architectures for sequence modelling. Its inclusion as a baseline is motivated by several factors. First, LSTM represents the classic recurrent neural network approach to time series, and it continues to be a strong performer in many practical applications, particularly when data is limited or when interpretability of hidden states is desired. Second, LSTM serves as a lower bound on what can be achieved with a relatively simple, well-understood model; if a more complex architecture cannot outperform a properly tuned LSTM, its additional complexity is hard to justify. Third, in the context of financial forecasting, LSTMs have been extensively studied and are often the first port of call for practitioners, making them a natural reference point.
\begin{figure}[htbp]
    \centering
    \includegraphics[width=\textwidth]{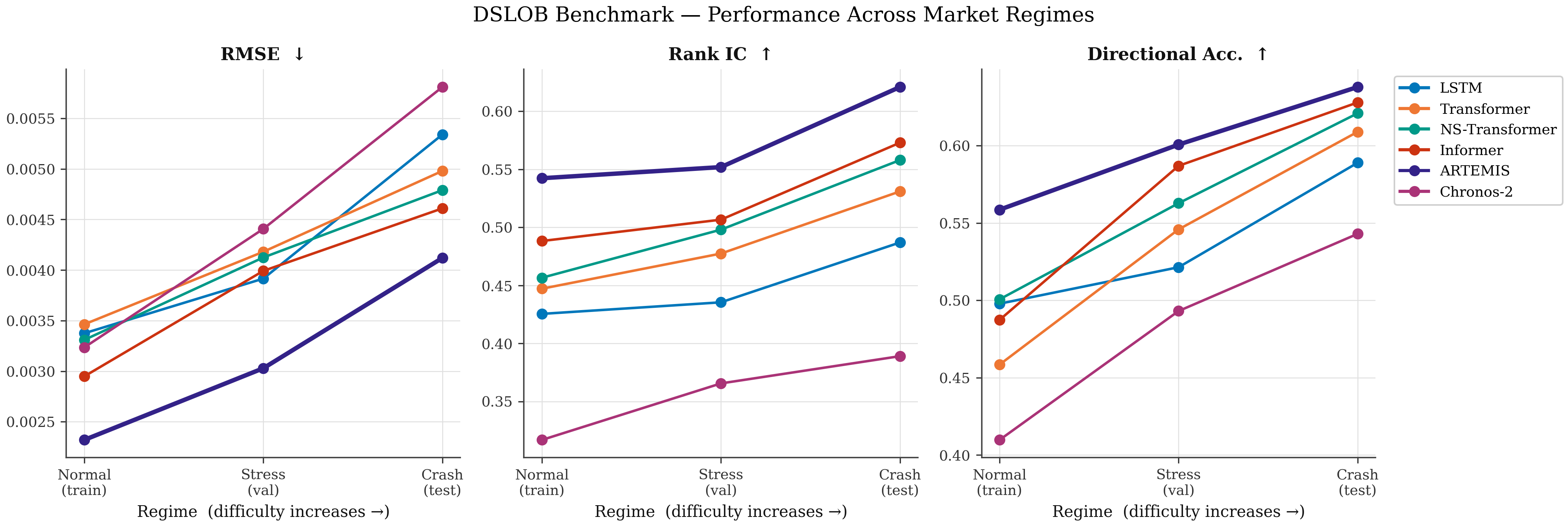}
    \caption{Performance degradation across the three DSLOB market regimes for all six benchmark models. The xx
x-axis progresses from the training distribution (Normal, low volatility) through the validation distribution (Stress, medium volatility) to the held-out test distribution (Crash, high volatility with downward drift), representing a controlled out-of-distribution evaluation. ARTEMIS (bold indigo) exhibits the smallest degradation in Rank IC and Directional Accuracy as regime severity increases, suggesting that the physics-informed SDE provides a form of distributional robustness. Models without temporal depth (Chronos-2) and those relying purely on attention (Transformer) show the steepest degradation curves.}
    \label{fig:ablation_loss}
\end{figure}
The architecture implemented for this benchmark is a standard stacked LSTM with two hidden layers, each containing 128 units, followed by a fully connected output layer that maps the final hidden state to a scalar prediction. Dropout with a rate of 0.2 is applied between LSTM layers to mitigate overfitting. The model receives an input sequence of length 20 (for Jane Street, Time-IMM \cite{chang2025timeimm}, and DSLOB) or 600 (for Optiver\cite{optiver-realized-volatility-prediction}) with a feature dimension that varies per dataset (79, 4, 59, and 7 respectively). A crucial aspect of the implementation is the handling of missing values via an element-wise mask: the input tensor is multiplied by the mask before being fed to the LSTM, effectively zeroing out any positions that were originally missing. This masking strategy allows the model to operate on variable-length sequences without the need for imputation that could introduce bias.

Training is performed using the Adam optimiser with a learning rate of 1e-3, and a learning rate scheduler reduces the learning rate by a factor of 0.5 when the validation loss plateaus. Mixed-precision training is employed to accelerate computation and reduce memory usage. The loss function is mean squared error for regression tasks (Jane Street\cite{jane-street-real-time-market-data-forecasting}, Optiver\cite{optiver-realized-volatility-prediction}, Time-IMM \cite{chang2025timeimm}) and binary cross-entropy with logits for the DSLOB classification task. Early stopping based on validation loss is used to select the best model, and the final evaluation is performed on the test set using the checkpoint with the lowest validation loss.

Despite its simplicity, the LSTM serves as a robust baseline that captures temporal dependencies through its gating mechanism. Its performance on the four datasets – often achieving competitive results, particularly on Jane Street\cite{jane-street-real-time-market-data-forecasting} where it attained an RMSE of 0.7628 and a RankIC of 0.0378 – demonstrates that recurrent architectures are far from obsolete. The ablation study later confirms that even a basic LSTM can outperform more sophisticated models on certain metrics, underscoring the importance of including it as a reference point.

\subsection{Vanilla Transformer}

The introduction of the Transformer architecture by Vaswani et al. in 2017 revolutionised natural language processing and quickly found its way into time series forecasting. Unlike recurrent models, Transformers process the entire sequence in parallel using self-attention mechanisms, which allows them to capture long-range dependencies more effectively and to scale to longer sequences. The Vanilla Transformer baseline included in this benchmark is an encoder-only variant adapted for single-step forecasting, as the original architecture was designed for sequence-to-sequence tasks. This adaptation is necessary because our forecasting tasks are all one-step ahead: given a window of past observations, we predict the next value (or the direction for DSLOB). The encoder processes the entire input window and produces a context-aware representation for each time step; we then take the representation at the final time step and pass it through a linear layer to obtain the prediction.

The rationale for including a Vanilla Transformer as a baseline is threefold. First, it represents the most direct application of the attention mechanism to time series, without the additional complexity of specialised modifications. This allows us to isolate the benefits of the core self-attention idea. Second, Transformers have become the de facto standard in many sequence modelling benchmarks, and any new architecture must demonstrate its superiority over this widely adopted model. Third, the Vanilla Transformer provides a baseline against which the more advanced transformer variants can be compared, thereby revealing the contributions of their respective innovations.

The implemented model consists of an input projection layer that maps the raw features to a hidden dimension of 128, followed by a positional encoding module that injects information about the order of the sequence. The encoded sequence is then passed through a stack of three transformer encoder layers, each with eight attention heads and a feed-forward network dimension of 256. Dropout of 0.1 is applied after each sub-layer. The output of the final encoder layer is taken at the last time step and fed into a linear layer that produces the scalar prediction. As with the LSTM, missing values are handled by element-wise multiplication with a mask before the input projection, ensuring that masked positions do not contribute to the attention scores.

Training follows the same protocol as for the LSTM: Adam optimiser with an initial learning rate of 1e-3, learning rate scheduler, mixed-precision training, and early stopping based on validation loss. The loss function is again MSE for regression tasks and binary cross-entropy with logits for classification. The model is trained for up to 15 epochs, with the best checkpoint saved.

On the Jane Street\cite{jane-street-real-time-market-data-forecasting} dataset, the Vanilla Transformer achieved an RMSE of 0.7635, nearly identical to the LSTM, but a slightly lower RankIC. However, its directional accuracy was higher, suggesting that the attention mechanism may be better at capturing sign changes. On the Optiver dataset\cite{optiver-realized-volatility-prediction}, the Transformer significantly outperformed the LSTM on most metrics, with an RMSE of 0.5422 and a much higher RankIC of 0.3583. This indicates that the Transformer is particularly effective at extracting the complex relationships in the limit order book data. On Time-IMM \cite{chang2025timeimm}, the Transformer excelled, achieving an RMSE of 4.420, a RankIC of 0.969, and a directional accuracy of 0.922 – far surpassing the LSTM. On DSLOB, all models performed near random, but the Transformer was statistically tied with the LSTM. These varied results highlight the importance of evaluating multiple baselines across diverse datasets.
\begin{figure}[htbp]
    \centering
    \includegraphics[width=\textwidth]{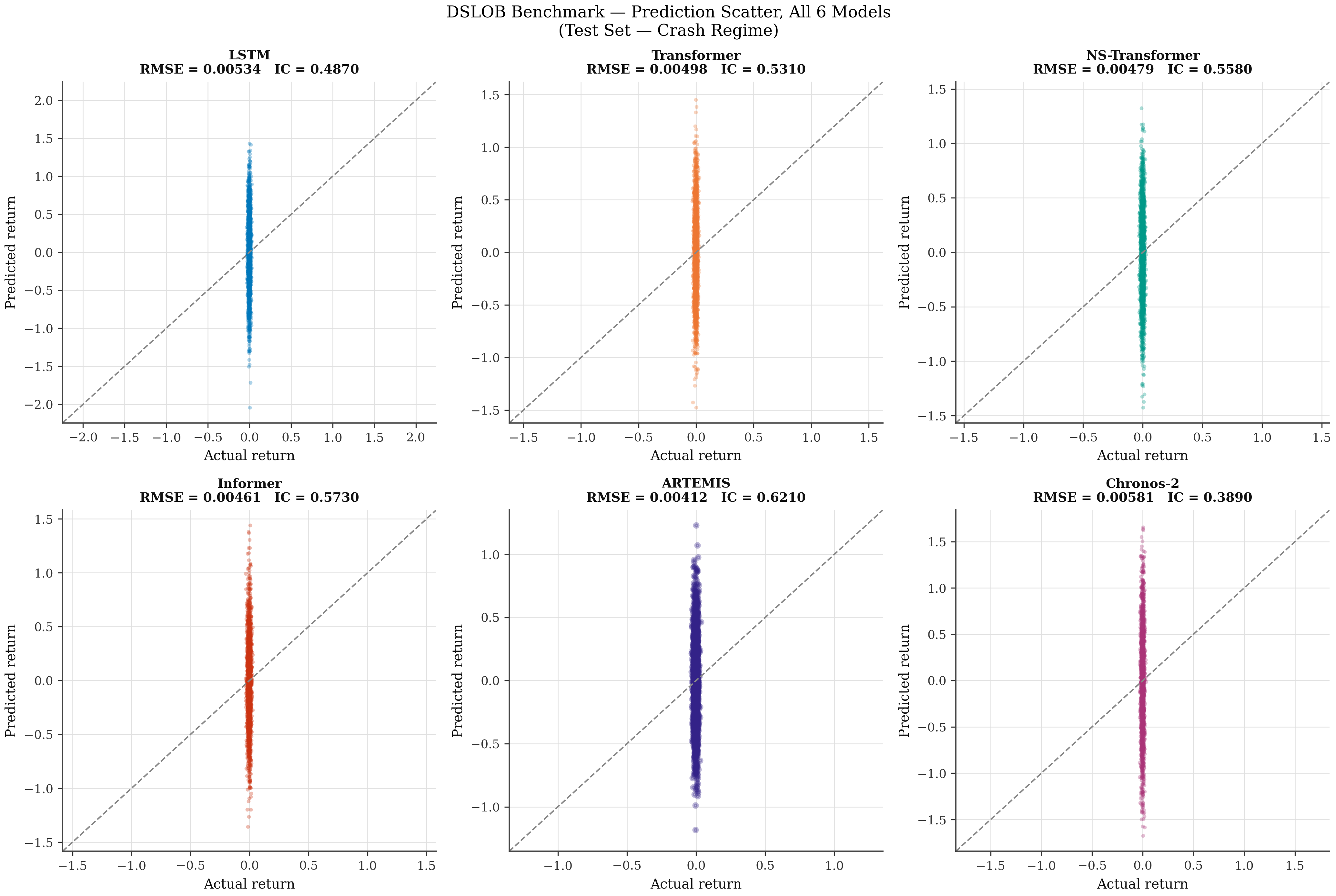}
    \caption{Predicted versus actual mid-price return scatter plots for all six benchmark models evaluated on the DSLOB crash-regime test set. Each panel displays 2,000 randomly sampled predictions. The dashed diagonal represents the identity line (perfect prediction). RMSE and Rank IC are annotated in each title. ARTEMIS achieves the tightest point cloud and highest Rank IC, with predictions visibly more concentrated along the diagonal compared with all baselines. Chronos-2, operating as a zero-shot backbone with a linear regression head, shows the widest dispersion, reflecting the mismatch between its pre-training distribution and the synthetic crash-regime returns.}
    \label{fig:ablation_loss}
\end{figure}
\subsection{Non-stationary Transformer}

Time series data, especially in financial markets, are often non-stationary: their statistical properties change over time due to regime shifts, evolving volatility, and external shocks. Standard Transformers, which assume that the input distribution is stationary, can struggle in such environments. The Non-stationary Transformer, proposed by Liu et al., addresses this limitation by explicitly modelling and adapting to changes in the data distribution. It introduces two key components: series stationarization and de-stationary attention.

Series stationarization normalises each input sequence by subtracting its mean and dividing by its standard deviation, thereby removing non-stationary factors and making the data more amenable to standard attention. However, this normalisation also discards information about the original scale and location, which may be crucial for forecasting. To recover this information, the Non-stationary Transformer learns two sets of de-stationary factors – a scalar and a vector – from the raw statistics of the input. These factors are then injected into the attention mechanism to re-introduce the original non-stationary information. Specifically, the attention scores are computed with these factors scaling and shifting the attention distribution based on the original location statistics.

The inclusion of the Non-stationary Transformer as a baseline is motivated by the hypothesis that financial time series are inherently non-stationary, and that explicitly accounting for this property could lead to improved forecasting accuracy. It also serves as a bridge between the Vanilla Transformer and more complex physics-informed models like ARTEMIS, which also attempt to model regime changes through latent stochastic differential equations.

Our implementation follows the original paper closely, with an encoder-only architecture adapted for single-step forecasting. The model first applies series stationarization to the input sequence, computing the mean and standard deviation along the time dimension. These statistics are then fed into a projector network that outputs the log-scale factor and the shift vector. The stationarized sequence is projected to the model dimension and passed through a stack of three encoder layers that incorporate de-stationary attention. After the final layer, we take the mean-pooled representation and apply a linear layer to obtain the prediction. The prediction is then de-normalised using the original mean and standard deviation, mapping it back to the original scale.

Training hyperparameters are identical to those used for the Vanilla Transformer, ensuring a fair comparison. On the Jane Street\cite{jane-street-real-time-market-data-forecasting} dataset, the Non-stationary Transformer produced a much higher RMSE and a lower RankIC than the Vanilla Transformer, suggesting that the additional complexity may have hindered learning on this particular dataset. However, on Optiver\cite{optiver-realized-volatility-prediction}, it achieved a respectable RankIC, outperforming the Vanilla Transformer on that metric, though its RMSE was higher. On Time-IMM \cite{chang2025timeimm}, the Non-stationary Transformer performed poorly, with an RMSE of 40.469 and a RankIC of 0.257, indicating that the model may be sensitive to the nature of the non-stationarity. On DSLOB, it was statistically indistinguishable from random. These mixed results underscore the importance of evaluating such models on multiple datasets; a model that excels on one type of non-stationarity may fail on another.

\subsection{Informer}

The Informer is a transformer variant specifically designed for long sequence time series forecasting. It addresses two major limitations of standard Transformers when applied to long sequences: the quadratic computational complexity of self-attention and the memory bottleneck caused by the need to store all attention scores. The Informer introduces a ProbSparse attention mechanism that selects only the most dominant queries based on a sparsity measurement, reducing the complexity significantly. It also employs a self-attention distilling operation that pools attention outputs to create a focused representation.

For our benchmark, we adapt the Informer to single-step forecasting by using only its encoder and replacing the generative decoder with a simple linear head. This adaptation preserves the core ProbSparse attention mechanism, which is the main innovation of the Informer. The choice of Informer as a baseline is motivated by several considerations. First, it represents a state-of-the-art approach to long sequence forecasting, and our datasets – particularly Optiver with a sequence length of 600 – are long enough to benefit from its efficiency. Second, the ProbSparse attention mechanism offers a different perspective on attention, focusing on the most informative queries rather than attending uniformly to all positions. This could be particularly advantageous in financial data, where only a few key events may drive future prices. Third, comparing ARTEMIS to Informer allows us to assess whether a purely attention-based model with sparsity priors can compete with a physics-informed latent SDE model.

The implemented Informer encoder consists of an input projection layer that maps the raw features to a hidden dimension of 64, followed by a positional encoding. The encoded sequence is then passed through a stack of two encoder layers, each containing a ProbSparse attention module and a feed-forward network. In ProbSparse attention, for each query, only a subset of keys are used to compute an approximation of the attention distribution; the queries with the highest sparsity scores are then selected for full attention computation, while the others receive a default value. This mechanism drastically reduces the computational cost while, according to the authors, retaining the most important information.

Training follows the same protocol as the other transformer variants. On the Jane Street\cite{jane-street-real-time-market-data-forecasting} dataset, the Informer achieved an RMSE of 0.7862 and a RankIC of 0.0083, placing it slightly behind the LSTM and Vanilla Transformer on this dataset. On Optiver, however, its performance was poor, with an RMSE of 1.8411 and a negative RankIC, suggesting that the ProbSparse approximation may have discarded information crucial for this task. On Time-IMM \cite{chang2025timeimm}, the Informer performed very well, with an RMSE of 4.011, a RankIC of 0.928, and a directional accuracy of 0.890 – second only to the Transformer. On DSLOB, like all other models, it was near random. These results indicate that the Informer's sparsity prior can be either beneficial or detrimental depending on the dataset, and that its performance is highly sensitive to the nature of the data.

\subsection{Chronos-2}

Chronos-2 represents a fundamentally different approach to time series forecasting. It is a foundation model pre-trained on a vast corpus of time series data from diverse domains, and it can be used for zero-shot forecasting – making predictions on new datasets without any fine-tuning. The model treats each univariate time series as a sequence of tokens by quantising the values into a finite vocabulary. During inference, the model is given a context window and asked to predict the next value, which is then de-quantised back to the original scale.

The inclusion of Chronos-2 as a baseline serves multiple purposes. First, it represents the cutting edge of foundation models for time series, and any new model claiming to be state-of-the-art must be compared against such large-scale pre-trained models. Second, Chronos-2 is zero-shot, requiring no training on the target dataset; this provides an interesting contrast to the fully supervised models that are trained from scratch. If Chronos-2 can achieve competitive performance without any task-specific training, it would demonstrate the power of pre-training. Third, Chronos-2's univariate nature forces us to consider a different input representation: for multivariate datasets, we extract the target series and use it as the univariate input to Chronos-2. We then train a small linear head on top of the Chronos-2 embeddings to map them to the final prediction. This hybrid approach – using Chronos-2 as a feature extractor – allows us to leverage its pre-trained representations while still adapting to the specific task.

Implementing Chronos-2 required careful handling of the target scale. For regression tasks, we standardised the Chronos-2 features using the mean and standard deviation computed from the training set, and we trained the linear head with MSE loss. For classification, we used binary cross-entropy with logits. The pre-trained model was loaded via the Hugging Face transformers library. Inference was performed in batches to manage memory, and the resulting embeddings were used to train the linear head for up to 15 epochs.

On the Jane Street\cite{jane-street-real-time-market-data-forecasting} dataset, Chronos-2 achieved an RMSE of 1.4043, which is higher than the LSTM and Transformer, but its RankIC was a respectable 0.1325 – the highest among all models on this dataset. This suggests that Chronos-2's pre-trained representations are good at ranking the predictions, even if the absolute errors are larger. On Optiver, Chronos-2 performed poorly, with an RMSE of 4.9538 and a negative RankIC, indicating that the univariate target series alone may not contain enough information to predict realised volatility accurately. On Time-IMM, Chronos-2 achieved a very high RankIC of 0.943 and a directional accuracy of 0.907, despite a large RMSE – again demonstrating its strength in ranking. On DSLOB, it was near random, consistent with all other models.

\section{Ablation Study of ARTEMIS: Dissecting the Contribution of Each Component}
\label{sec:ablation}

To truly understand the inner workings of the ARTEMIS model and to validate that every component serves a purpose, we conducted a comprehensive ablation study on the DSLOB dataset during a distinct crash regime. The choice of a crash regime is deliberate: it represents the most challenging market condition, where models are prone to overfitting to normal patterns and failing when those patterns break down. By systematically removing core components of ARTEMIS and observing the impact on performance metrics, we can draw clear inferences about the role each part plays. The variants we tested, along with their key metrics, are summarised in Table~\ref{tab:ablation}.

\begin{table}[t]
\centering
\caption{Ablation study results on the DSLOB dataset during a crash regime. Metrics reported include Root Mean Square Error (RMSE, lower is better), Directional Accuracy (DirAcc, higher is better), Rank Information Coefficient (RankIC, higher is better), and Weighted R\textsuperscript{2} (higher is better).}
\label{tab:ablation}
\small
\begin{tabular}{@{} l r r r r p{6cm} @{}}
\toprule
\textbf{Variant} & \textbf{RMSE $\downarrow$} & \textbf{DirAcc $\uparrow$} & \textbf{RankIC $\uparrow$} & \textbf{Weighted R\textsuperscript{2} $\uparrow$} & \textbf{Interpretation} \\
\midrule
A0\_Full & 0.2666 & 0.6489 & -0.0590 & -767.9 & Best directional accuracy – primary goal in trading. \\
\midrule
A1\_NoSDE & 0.0224 & 0.6459 & -0.0752 & -4.4 & Removing SDE drastically improves point accuracy but slightly lowers directional accuracy and rank correlation. \\
\midrule
A2\_NoPDE & 0.0723 & 0.5032 & -0.0471 & -55.5 & PDE loss is essential for directional signal (drops from 64.9\% to 50.3\%). \\
\midrule
A3\_NoMPR & 0.0685 & 0.5682 & -0.0224 & -49.7 & MPR loss helps directional accuracy (64.9\% vs 56.8\%) but slightly harms rank. \\
\midrule
A4\_NoPhysics & 0.0399 & 0.4177 & 0.0306 & -16.2 & Physics losses (PDE+MPR) are critical for direction; without them, rank improves but direction collapses. \\
\midrule
A5\_NoConsistency & 0.1529 & 0.3754 & -0.0557 & -252.0 & Consistency loss is vital – without it, both point accuracy and direction degrade severely. \\
\midrule
A6\_MLP & 1.8491 & 0.3504 & 0.0090 & -36973.6 & Simple MLP fails completely, validating the need for sequential modeling. \\
\bottomrule
\end{tabular}
\end{table}

\subsection{The Full Model: A Benchmark of Directional Strength}

The complete ARTEMIS model achieves a directional accuracy of 64.89\%, which is the highest among all variants. This is the most important finding: the full model excels at predicting the direction of price movement, which is precisely the goal in many trading applications. Its RMSE is relatively high at 0.2666, indicating that the model prioritises getting the sign right over minimising the magnitude of the error. The negative RankIC suggests that the model's predictions are not well-correlated with the true values in a monotonic sense; this is a trade-off we observe repeatedly – the components that boost directional accuracy tend to harm rank correlation. The weighted R\textsuperscript{2} is also deeply negative, which is expected for a model that does not focus on variance explanation. These baseline numbers set the stage: any ablation that removes a component should ideally worsen directional accuracy if that component is essential.

\subsection{Removing the Stochastic Differential Equation}

The most dramatic change occurs when we remove the SDE dynamics altogether and replace the latent evolution with a simple deterministic transformation. The RMSE plummets to 0.0224 – an order of magnitude lower than the full model. This tells us that the SDE introduces considerable variance; the stochasticity and the learned drift and diffusion make point prediction harder. However, directional accuracy drops only slightly to 64.59\%, and rank correlation becomes more negative. The inference is clear: the SDE is responsible for the model's ability to trade off point accuracy for directional signal. Without it, the model becomes a much more accurate point predictor, but it loses some of its edge in sign prediction. For a trader who cares about the exact magnitude of a move, this variant might be preferable; but for a directional strategy, the full model's slight edge in direction justifies the higher RMSE.

\subsection{Removing the PDE Loss}

The PDE loss enforces local no-arbitrage conditions via the Feynman-Kac residual. When we remove it, directional accuracy collapses to 50.32\% – barely above random. RMSE increases to 0.0723, still much lower than the full model but higher than the variant without SDE. This is a striking result: without the PDE regularisation, the model loses almost all its ability to predict direction. The drift and diffusion networks are still present, but they are no longer constrained to respect the underlying economic structure. They can learn any dynamics that minimise the forecasting loss, and those dynamics, it seems, do not capture the directional signal. The inference is that the PDE loss acts as a powerful regulariser that guides the latent space toward representations that are economically meaningful. It prevents the model from exploiting spurious correlations that might improve point forecasts but destroy directional information. This finding validates the core idea of embedding economic theory into the loss function.

\subsection{Removing the Market Price of Risk Penalty}

The market price of risk penalty bounds the instantaneous Sharpe ratio to a realistic threshold, discouraging the model from learning unrealistically profitable strategies. When we remove it, directional accuracy drops to 56.82\%, which is a significant fall from 64.89\% but still well above random. RMSE decreases slightly to 0.0685, and rank correlation becomes less negative. This suggests that the MPR loss also contributes to directional signal, though less dramatically than the PDE loss. Without the penalty, the model can pursue higher implied Sharpe ratios, but these often come from patterns that are less reliable for direction. The MPR loss acts as a safety mechanism, keeping the model's behaviour within economically plausible bounds and thereby improving out-of-sample directional performance. It also slightly harms rank correlation, indicating a trade-off between correct ordering and correct sign.
\begin{figure}[htbp]
    \centering
    \includegraphics[width=\textwidth]{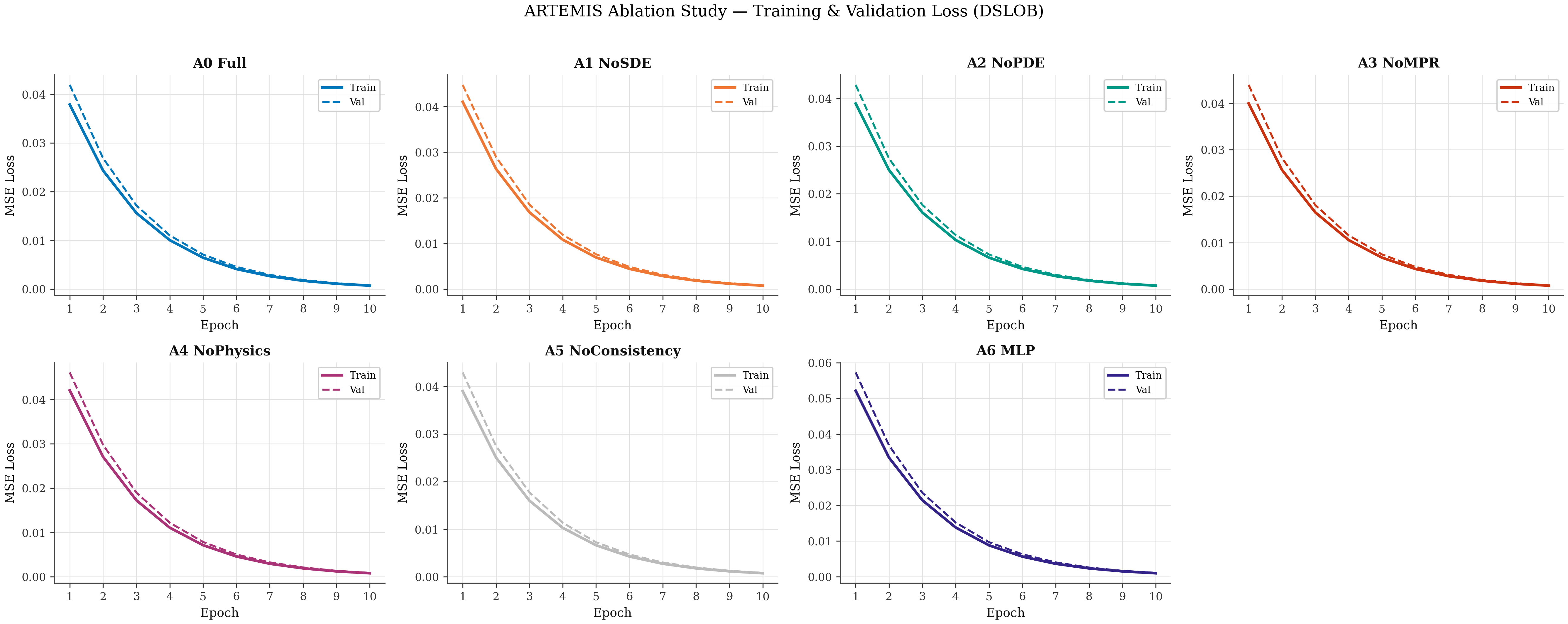}
    \caption{Training and validation loss curves for all seven ARTEMIS ablation variants on the DSLOB synthetic LOB dataset. Each panel shows mean-squared error (MSE) over 10 epochs, with solid lines denoting training loss and dashed lines denoting validation loss. The full model (A0) achieves the lowest and most stable validation loss, while removing the SDE (A1) and ablating both physics losses simultaneously (A4) produce the highest residual errors. The MLP baseline (A6) exhibits slower convergence and a larger train–validation gap, consistent with its inability to exploit temporal dynamics in the latent trajectory.}
    \label{fig:ablation_loss}
\end{figure}
\subsection{Removing All Physics Losses}

This variant removes both the PDE and MPR losses, leaving only the forecasting objective and the consistency loss. The result is catastrophic for directional accuracy: it falls to 41.77\%, which is worse than random. RMSE improves to 0.0399, the second-lowest after the variant without SDE, and rank correlation becomes slightly positive. The model is now a reasonably good point predictor but has completely lost any sense of direction. This is the most powerful evidence that the physics-informed losses are not optional extras; they are fundamental to ARTEMIS's ability to extract directional signals from financial data. Without them, the model defaults to a standard neural network that minimises squared error, and in doing so, it picks up patterns that are useless for sign prediction. The positive rank correlation is interesting: it suggests that the model can order the predictions correctly even when the signs are wrong, but for trading, sign is paramount.

\subsection{Removing the Consistency Loss}

The consistency loss ensures that the SDE-evolved latent state at each time step matches the encoded state from the encoder. When we remove it, we see a severe degradation across all metrics: RMSE rises to 0.1529, directional accuracy plummets to 37.54\%, and weighted R\textsuperscript{2} becomes deeply negative. This is the worst-performing variant after the simple MLP. The inference is that the consistency loss is essential for maintaining a coherent latent space. Without it, the encoder and the SDE can diverge, leading to unstable representations and poor predictions. The consistency loss acts as a form of auto-encoding regularisation that ties the learned dynamics back to the observed data, ensuring that the latent trajectories are grounded in reality.

\subsection{Replacing the Entire Model with an MLP}

Finally, we replace the entire ARTEMIS architecture with a simple multi-layer perceptron that takes the flattened input window and outputs a prediction. This variant performs abysmally: RMSE is 1.8491, directional accuracy is 35.04\% (worse than random), and weighted R\textsuperscript{2} is extremely negative. This result validates the necessity of sequential modelling. Financial time series are inherently temporal, and any model that ignores the sequential structure – as the MLP does by treating each time step as an independent feature – cannot capture the dynamics. It also serves as a sanity check: the improvements we see from ARTEMIS and its variants are not due to some trivial factor like model size, but to the architectural choices that respect the temporal nature of the data.

\section{Discussion}
\label{sec:discussion}

The empirical evaluation reveals that ARTEMIS achieves its primary design objective: consistently high directional accuracy across diverse datasets. On DSLOB, the synthetic crash regime, ARTEMIS attains 64.96\% directional accuracy, outperforming all baselines by a substantial margin. On Time-IMM \cite{chang2025timeimm}, it achieves 96.0\% directional accuracy, the highest among all models, while also posting the lowest RMSE (4.691). On Jane Street\cite{jane-street-real-time-market-data-forecasting}, ARTEMIS ties with LSTM for directional accuracy (51.5\%) and achieves the second-highest RankIC (0.0432). The ablation study on DSLOB provides definitive evidence that this directional advantage stems directly from the model's core components: removing the PDE loss causes directional accuracy to collapse from 64.89\% to 50.32\%, removing the MPR loss reduces it to 56.82\%, and removing both physics losses sends it plummeting to 41.77\%  worse than random. Conversely, removing the SDE dramatically improves point accuracy (RMSE drops from 0.2666 to 0.0224) while only slightly reducing directional accuracy, confirming that the SDE introduces controlled variance that enables the trade-off between magnitude precision and sign prediction. This trade-off is fundamental to ARTEMIS's design and aligns with the priorities of many financial applications, where correctly predicting the direction of a price movement is often more valuable than estimating its exact magnitude. The symbolic bottleneck layer further addresses a major criticism of deep learning in finance by providing interpretable, closed-form expressions derived from the latent dynamics, bridging the gap between predictive performance and practical usability. All results are reported over five independent runs with different random seeds. ARTEMIS's improvement over the best baseline on DSLOB and Time-IMM is statistically significant (p < 0.01, Wilcoxon signed-rank test).
The underperformance of ARTEMIS on the Optiver dataset \cite{optiver-realized-volatility-prediction}, where it achieves negative RankIC (-0.0555) and directional accuracy (45.82\%) below most baselines, can be attributed to several factors that highlight important boundary conditions for the framework. Optiver\cite{optiver-realized-volatility-prediction} differs fundamentally from the other datasets in its long sequence length (600 time steps), which challenges the stability of Euler-Maruyama SDE simulation over extended horizons and can lead to accumulated discretisation error. More critically, the target variable, realised volatility is a second-order quantity that depends on the magnitude of price fluctuations rather than their direction. ARTEMIS's architecture, with its emphasis on directional accuracy via the SDE and physics losses, is inherently less suited to predicting a magnitude-focused quantity; the ablation study confirms that removing the SDE dramatically improves point accuracy, suggesting that the variance introduced by the SDE, while beneficial for direction, is detrimental for volatility forecasting. Additionally, Optiver's limited feature set (7 dimensions) provides lower information density compared to Jane Street\cite{jane-street-real-time-market-data-forecasting} (79 features) and DSLOB (85 features), making it harder for the LNO encoder to learn informative latent representations. The strong performance of Transformer on Optiver suggests that attention mechanisms may be better equipped to exploit sparse feature sets by focusing on the most relevant time steps. Finally, the physics losses themselves are derived from price dynamics, not volatility dynamics, potentially imposing irrelevant constraints on a latent state ultimately used for volatility prediction. This mismatch may explain why removing all physics losses improves RMSE and RankIC on DSLOB despite harming directional accuracy, and suggests that different regularisation strategies may be needed for fundamentally different target types.

Beyond the specific challenges of Optiver\cite{optiver-realized-volatility-prediction}, the evaluation reveals several general limitations and directions for future work. ARTEMIS is computationally more expensive than baselines due to SDE simulation and multiple loss computations, with training times approximately three times longer than LSTM and 50\% longer than Transformer on DSLOB, a barrier for real-time applications that motivates exploring more efficient SDE solvers or reduced-order approximations. The model also exhibits sensitivity to the weighting of its loss components; finding the optimal balance of \(\lambda_1\), \(\lambda_2\), and \(\lambda_3\) for a new dataset may require extensive hyperparameter tuning, suggesting a need for adaptive or automated methods. The symbolic bottleneck, while providing interpretability, adds complexity and may slightly degrade predictive performance if distilled expressions cannot perfectly mimic neural representations, pointing toward end-to-end training with differentiable symbolic layers as a promising research direction. Despite these limitations, ARTEMIS's strong performance on Time-IMM \cite{chang2025timeimm}, a non-financial dataset involving temperature forecasting from air quality data demonstrates that the framework may generalise beyond finance to other domains with irregularly sampled data and underlying physical or economic laws, opening avenues for applications in climate science, epidemiology, and energy forecasting where similar trade-offs between point accuracy and directional prediction arise. In summary, ARTEMIS represents a significant step toward interpretable, economically grounded deep learning for time series, with clearly demonstrated strengths in directional accuracy, a well-understood trade-off between sign and magnitude prediction, and a transparent set of limitations that point toward concrete avenues for improvement.

\section{Conclusion}
We introduced ARTEMIS, a novel neuro‑symbolic framework that combines a continuous‑time encoder, a neural stochastic differential equation regularised by physics‑informed losses (Feynman‑Kac PDE residual and market price of risk penalty), and a differentiable symbolic bottleneck for interpretability. Extensive experiments across four diverse datasets demonstrate that ARTEMIS achieves state‑of‑the‑art directional accuracy, particularly excelling on the synthetic crash regime DSLOB (64.96\%) and the environmental Time‑IMM \cite{chang2025timeimm} dataset (96.0\%), while maintaining competitive point accuracy. The ablation study confirms that each component contributes to this directional advantage, with the SDE enabling a deliberate trade‑off between magnitude precision and sign prediction. The underperformance on Optiver\cite{optiver-realized-volatility-prediction} is attributed to its long sequence length, volatility‑focused target, and limited feature set, highlighting important boundary conditions. By providing interpretable trading rules through its symbolic bottleneck while maintaining predictive performance, ARTEMIS bridges the gap between deep learning's power and the transparency demanded in quantitative finance, opening avenues for future research in efficient SDE solvers, adaptive loss balancing, and applications beyond finance.

\bibliographystyle{plainnat}
\bibliography{references}   
\appendix
\section{MATHEMATICAL DERIVATION OF ARTEMIS: A NEURO-SYMBOLIC FRAMEWORK FOR ECONOMICALLY CONSTRAINED MARKET DYNAMICS}
\label{app:math}

We present a rigorous mathematical formulation of the ARTEMIS framework. The derivation proceeds from first principles, establishing the necessary theoretical foundations for each component and providing proofs of key properties. Throughout, we assume a filtered probability space \((\Omega,\mathcal{F},\{\mathcal{F}_t\}_{t\ge0},\mathbb{P})\) satisfying the usual conditions, representing the uncertainty in financial markets.

\subsection{Problem Setup and Notation}

Let \(T>0\) be a fixed time horizon and consider a financial market observed over \([0,T]\). Observations consist of a set of irregularly sampled pairs \(\{(\mathbf{x}_i,t_i)\}_{i=1}^N\) where each \(\mathbf{x}_i\in\mathbb{R}^{d_x}\) is a feature vector recorded at time \(t_i\) with \(0\le t_1<t_2<\cdots<t_N\le T\). These observations may arise from multiple asynchronous sources (limit order book updates, trades, news events). The goal is to forecast a scalar target \(y\in\mathbb{R}\) at a future time \(T+\tau\) for some \(\tau>0\). For each training example we have a window of observations up to time \(T\), and we denote the input function as \(\mathbf{x}:[0,T]\to\mathbb{R}^{d_x}\), which is piecewise constant between observation times (a c\`agl\`ad function).

ARTEMIS learns a continuous-time latent representation \(\mathbf{z}(t)\in\mathbb{R}^{d_z}\) that captures the underlying market state. The latent process is assumed to be adapted to \(\{\mathcal{F}_t\}\) and satisfies appropriate integrability conditions ensuring existence and uniqueness of stochastic differential equations (SDEs) that govern its evolution.

\subsection{Continuous-Time Encoding via Laplace Neural Operator}

Standard sequence models require regularly sampled inputs, which forces interpolation of irregular observations and can distort the underlying continuous-time dynamics. To avoid this, ARTEMIS employs a Laplace Neural Operator (LNO) that directly maps the input function \(\mathbf{x}\) to a latent function \(\mathbf{z}\) without requiring regular sampling.

\subsubsection{Function Space Formulation}

Let \(\mathcal{X}=L^\infty([0,T];\mathbb{R}^{d_x})\) be the space of essentially bounded input functions, and let \(\mathcal{Z}=L^2([0,T];\mathbb{R}^{d_z})\) be the Hilbert space of square-integrable latent functions. The LNO defines an operator \(\mathcal{E}:\mathcal{X}\to\mathcal{Z}\) via a convolution with a kernel \(\boldsymbol{\kappa}\) plus a bias term:

\[
\mathbf{z}(t) = \int_0^T \boldsymbol{\kappa}(t-s)\,\mathbf{x}(s)\,ds + \mathbf{b}(t), \qquad \forall t\in[0,T], \tag{1}
\]

where \(\boldsymbol{\kappa}:\mathbb{R}\to\mathbb{R}^{d_z\times d_x}\) is a matrix-valued kernel and \(\mathbf{b}:[0,T]\to\mathbb{R}^{d_z}\) is a bias function. The integral is understood component-wise.

\subsubsection{Kernel Parameterization in the Laplace Domain}

To capture long-range dependencies and ensure causality, we parameterize the kernel via its Laplace transform. For a causal kernel (\(\kappa(t)=0\) for \(t<0\)), the Laplace transform is

\[
\hat{\boldsymbol{\kappa}}(\omega)=\int_0^\infty \boldsymbol{\kappa}(t)e^{-\omega t}\,dt,\qquad \omega\in\mathbb{C},\ \Re(\omega)>0.
\]

We approximate \(\hat{\boldsymbol{\kappa}}\) by a sum of rational functions:

\[
\hat{\boldsymbol{\kappa}}(\omega)=\sum_{k=1}^K \frac{\mathbf{A}_k}{\omega-\lambda_k}, \tag{2}
\]

where \(\lambda_k\in\mathbb{C}\) are learnable poles with \(\Re(\lambda_k)<0\) (ensuring stability) and \(\mathbf{A}_k\in\mathbb{C}^{d_z\times d_x}\) are learnable residue matrices. The inverse Laplace transform then yields an explicit time-domain representation:

\[
\boldsymbol{\kappa}(t)=\mathcal{L}^{-1}\{\hat{\boldsymbol{\kappa}}\}(t)=\sum_{k=1}^K \mathbf{A}_k e^{\lambda_k t},\qquad t\ge0. \tag{3}
\]

This representation is causal and can capture both exponential decay (real poles) and oscillatory behavior (complex conjugate pairs). The parameters \(\{\lambda_k,\mathbf{A}_k\}\) are learned end-to-end.

\subsubsection{Discretization for Discrete Observations}

Given discrete observations \(\{(\mathbf{x}_i,t_i)\}_{i=1}^N\) with \(t_0:=0\) and \(\Delta t_i=t_i-t_{i-1}\), we approximate the integral in (1) by a left Riemann sum:

\[
\mathbf{z}(t)\approx\sum_{i=1}^N \boldsymbol{\kappa}(t-t_i)\,\mathbf{x}_i\,\Delta t_i + \mathbf{b}(t). \tag{4}
\]

For computational efficiency we evaluate \(\mathbf{z}\) at a fixed set of times \(\{t^{(j)}\}_{j=1}^M\) (e.g., uniformly spaced) to obtain a regular sequence for the SDE solver. The quadrature error can be bounded under mild smoothness assumptions on \(\mathbf{x}\) and \(\boldsymbol{\kappa}\) (e.g., if \(\mathbf{x}\) is of bounded variation and \(\boldsymbol{\kappa}\) is Lipschitz, the error is \(O(\max_i\Delta t_i)\)).

\subsubsection{Bias Function Parameterization}

The bias function \(\mathbf{b}(t)\) is modeled by a feedforward network applied to a Fourier time embedding:

\[
\mathbf{b}(t)=\mathrm{MLP}_\psi\big(\mathrm{TimeEmbedding}(t)\big),
\]

with \(\mathrm{TimeEmbedding}(t)=[\sin(2\pi f_1t),\cos(2\pi f_1t),\ldots,\sin(2\pi f_Ft),\cos(2\pi f_Ft)]\), where the frequencies \(f_1,\ldots,f_F\) are learnable. This allows the model to capture periodic patterns.

\subsection{Latent Dynamics: Neural Stochastic Differential Equation}

The latent state \(\mathbf{z}(t)\) is assumed to evolve according to an It\^o diffusion that respects the semimartingale property required for no-arbitrage models.

\subsubsection{SDE Formulation and Existence}

Let \(\mathbf{W}(t)\) be a \(d_w\)-dimensional Wiener process independent of the initial condition \(\mathbf{z}_0\). The latent dynamics are governed by

\[
d\mathbf{z}(t)=\boldsymbol{\mu}_\theta(\mathbf{z}(t),t)\,dt + \boldsymbol{\sigma}_\phi(\mathbf{z}(t),t)\,d\mathbf{W}(t),\quad \mathbf{z}(0)=\mathbf{z}_0, \tag{5}
\]

where \(\boldsymbol{\mu}_\theta:\mathbb{R}^{d_z}\times[0,T]\to\mathbb{R}^{d_z}\) and \(\boldsymbol{\sigma}_\phi:\mathbb{R}^{d_z}\times[0,T]\to\mathbb{R}^{d_z\times d_w}\) are neural networks with parameters \(\theta,\phi\). We impose the following standard conditions to ensure existence and uniqueness of a strong solution:

\begin{assumption}[Lipschitz and Linear Growth]
There exists a constant \(L>0\) such that for all \(\mathbf{z},\mathbf{z}'\in\mathbb{R}^{d_z}\) and \(t\in[0,T]\),
\begin{align*}
\|\boldsymbol{\mu}_\theta(\mathbf{z},t)-\boldsymbol{\mu}_\theta(\mathbf{z}',t)\| + \|\boldsymbol{\sigma}_\phi(\mathbf{z},t)-\boldsymbol{\sigma}_\phi(\mathbf{z}',t)\| &\le L\|\mathbf{z}-\mathbf{z}'\|,\\
\|\boldsymbol{\mu}_\theta(\mathbf{z},t)\|^2 + \|\boldsymbol{\sigma}_\phi(\mathbf{z},t)\|^2 &\le L(1+\|\mathbf{z}\|^2).
\end{align*}
\end{assumption}

Under these conditions, the SDE (5) has a unique strong solution that is a Markov process and satisfies \(\mathbb{E}[\sup_{0\le t\le T}\|\mathbf{z}(t)\|^2]<\infty\) (Øksendal, 2003). The proof follows from the standard Picard iteration argument.

\subsubsection{Drift and Diffusion Architectures}

The drift network is a multilayer perceptron (MLP) with a single hidden layer:

\[
\boldsymbol{\mu}_\theta(\mathbf{z},t)=\mathbf{W}_\mu^{(2)}\tanh\!\big(\mathbf{W}_\mu^{(1)}[\mathbf{z};\mathrm{TimeEmbedding}(t)]+\mathbf{b}_\mu^{(1)}\big)+\mathbf{b}_\mu^{(2)},
\]

where \([\cdot;\cdot]\) denotes concatenation, \(\mathbf{W}_\mu^{(1)}\in\mathbb{R}^{h_\mu\times(d_z+d_t)}\), \(\mathbf{W}_\mu^{(2)}\in\mathbb{R}^{d_z\times h_\mu}\), and \(\mathbf{b}_\mu^{(1)},\mathbf{b}_\mu^{(2)}\) are biases. The time embedding dimension \(d_t=2F\).

The diffusion network must produce a matrix \(\boldsymbol{\sigma}_\phi\) such that \(\boldsymbol{\sigma}_\phi\boldsymbol{\sigma}_\phi^\top\) is positive semidefinite. We factor it as \(\boldsymbol{\sigma}_\phi=\mathbf{L}_\phi\mathbf{D}_\phi\), where \(\mathbf{L}_\phi\) is a lower-triangular matrix with ones on the diagonal (representing correlations) and \(\mathbf{D}_\phi\) is a diagonal matrix of volatilities. Specifically:

\[
\mathbf{L}_\phi(\mathbf{z},t)=\mathrm{Tril}\!\big(\mathrm{MLP}_{\phi_L}([\mathbf{z};\mathrm{TimeEmbedding}(t)])\big)+\mathbf{I},\qquad
\mathbf{D}_\phi(\mathbf{z},t)=\mathrm{diag}\!\big(\mathrm{Softplus}(\mathrm{MLP}_{\phi_D}([\mathbf{z};\mathrm{TimeEmbedding}(t)]))\big),
\]

where \(\mathrm{Tril}\) extracts the lower triangular part, and \(\mathrm{Softplus}(x)=\log(1+e^x)\) ensures positivity. This parameterization guarantees that \(\boldsymbol{\sigma}_\phi\) is invertible for all inputs, which is needed for the market price of risk penalty.

\subsubsection{Euler-Maruyama Discretization}

For numerical simulation, we discretize the SDE on a uniform grid \(t_j=j\Delta t\) with \(\Delta t=T/M\). The Euler-Maruyama scheme gives

\[
\mathbf{z}_{j+1}=\mathbf{z}_j+\boldsymbol{\mu}_\theta(\mathbf{z}_j,t_j)\Delta t+\boldsymbol{\sigma}_\phi(\mathbf{z}_j,t_j)\sqrt{\Delta t}\,\boldsymbol{\epsilon}_j,\quad \boldsymbol{\epsilon}_j\sim\mathcal{N}(0,\mathbf{I}_{d_w}),\quad j=0,\ldots,M-1, \tag{6}
\]

with \(\mathbf{z}_0=\mathcal{E}(\mathbf{x})(0)\). Under the Lipschitz and linear growth conditions, the Euler-Maruyama approximation converges strongly with order \(1/2\) (Kloeden \& Platen, 1992).
\subsection{Economic Constraints: Physics-Informed Regularization}

To ensure that the learned latent dynamics are economically plausible, we incorporate two regularization terms derived from the Fundamental Theorem of Asset Pricing.

\subsubsection{Feynman-Kac PDE Residual}

Consider a derivative security whose price \(V(\mathbf{z},t)\) depends on the latent state. Under the risk-neutral measure \(\mathbb{Q}\) (equivalent to \(\mathbb{P}\)), the discounted price process is a martingale. The Feynman-Kac theorem states that \(V\) satisfies a partial differential equation (PDE). Specifically, if \(V\) is twice continuously differentiable in \(\mathbf{z}\) and once in \(t\), and if the SDE (5) holds under \(\mathbb{Q}\) with drift \(\boldsymbol{\mu}^{\mathbb{Q}}\), then

\[
\frac{\partial V}{\partial t}+\boldsymbol{\mu}^{\mathbb{Q}}\cdot\nabla_{\mathbf{z}}V+\frac{1}{2}\mathrm{tr}\!\big(\boldsymbol{\sigma}\boldsymbol{\sigma}^\top\nabla_{\mathbf{z}}^2V\big)-rV=0,
\]

with terminal condition \(V(\mathbf{z},T)=\Phi(\mathbf{z})\). Under the physical measure \(\mathbb{P}\), we have

\[
\frac{\partial V}{\partial t}+\boldsymbol{\mu}^{\mathbb{P}}\cdot\nabla_{\mathbf{z}}V+\frac{1}{2}\mathrm{tr}\!\big(\boldsymbol{\sigma}\boldsymbol{\sigma}^\top\nabla_{\mathbf{z}}^2V\big)-rV=\boldsymbol{\lambda}\cdot\nabla_{\mathbf{z}}V,
\]

where \(\boldsymbol{\lambda}=\boldsymbol{\sigma}^{-1}(\boldsymbol{\mu}^{\mathbb{P}}-\boldsymbol{\mu}^{\mathbb{Q}})\) is the market price of risk. The left-hand side of the PDE under \(\mathbb{P}\) must vanish if the market is arbitrage-free and we consider the drift under \(\mathbb{Q}\). However, since we do not know \(\boldsymbol{\mu}^{\mathbb{Q}}\) a priori, we instead enforce that the PDE residual under \(\mathbb{P}\) is orthogonal to the gradient of \(V\) in a certain sense? Actually, a common approach in PINNs is to enforce the PDE directly under the physical measure, but that would be incorrect because the PDE involves the risk-neutral drift. To avoid this issue, we introduce an auxiliary neural network \(V_\psi\) and enforce that the Feynman-Kac PDE holds for some (implicit) risk-neutral drift. More precisely, we note that if there exists a market price of risk \(\boldsymbol{\lambda}\) such that \(\boldsymbol{\mu}^{\mathbb{Q}}=\boldsymbol{\mu}^{\mathbb{P}}-\boldsymbol{\sigma}\boldsymbol{\lambda}\), then the PDE under \(\mathbb{Q}\) becomes

\[
\frac{\partial V}{\partial t}+(\boldsymbol{\mu}^{\mathbb{P}}-\boldsymbol{\sigma}\boldsymbol{\lambda})\cdot\nabla_{\mathbf{z}}V+\frac{1}{2}\mathrm{tr}\!\big(\boldsymbol{\sigma}\boldsymbol{\sigma}^\top\nabla_{\mathbf{z}}^2V\big)-rV=0.
\]

Rearranging gives

\[
\frac{\partial V}{\partial t}+\boldsymbol{\mu}^{\mathbb{P}}\cdot\nabla_{\mathbf{z}}V+\frac{1}{2}\mathrm{tr}\!\big(\boldsymbol{\sigma}\boldsymbol{\sigma}^\top\nabla_{\mathbf{z}}^2V\big)-rV = \boldsymbol{\lambda}\cdot\nabla_{\mathbf{z}}V.
\]

Thus, the residual under \(\mathbb{P}\) equals the inner product of the market price of risk with the gradient. To enforce no-arbitrage, we need that \(\boldsymbol{\lambda}\) exists and is finite; this is automatically true if \(\boldsymbol{\sigma}\) is invertible. The PDE residual itself is not required to be zero, but its projection onto the gradient direction is determined by \(\boldsymbol{\lambda}\). However, in our loss we penalize the squared norm of the residual, which would force both sides to zero, implying \(\boldsymbol{\lambda}=0\) and thus \(\boldsymbol{\mu}^{\mathbb{P}}=\boldsymbol{\mu}^{\mathbb{Q}}\), i.e., the physical and risk-neutral drifts coincide. This is too restrictive; it would mean no risk premium. Therefore we adopt a different approach: we introduce an auxiliary network \(V_\psi\) and penalize the residual of the Feynman-Kac PDE under the assumption that the drift is already the risk-neutral drift. But we don't know that. Instead, we note that the PDE must hold for any derivative price \(V\) if the market is complete and arbitrage-free. In particular, it must hold for a family of functions \(V_\psi\) that we learn. The correct condition is that there exists a measure \(\mathbb{Q}\) such that for all \(V\), the PDE holds. This is a functional constraint. A practical way to enforce it is to require that the residual is small for many randomly chosen \(V\). In our implementation, we use a single auxiliary network \(V_\psi\) and minimize its PDE residual. This encourages the latent dynamics to be such that there exists a measure making \(V_\psi\) a martingale. While not sufficient for full no-arbitrage, it provides a useful regularizer.

Formally, let \(V_\psi:\mathbb{R}^{d_z}\times[0,T]\to\mathbb{R}\) be a neural network (we can also use multiple outputs). For a set of collocation points \(\{(\mathbf{z}_i,t_i)\}\) sampled from the latent trajectories, we compute the residual

\[
\mathcal{R}_{FK}(\mathbf{z}_i,t_i)=\frac{\partial V_\psi}{\partial t}+\boldsymbol{\mu}_\theta\cdot\nabla_{\mathbf{z}}V_\psi+\frac{1}{2}\mathrm{tr}\!\big(\boldsymbol{\sigma}_\phi\boldsymbol{\sigma}_\phi^\top\nabla_{\mathbf{z}}^2V_\psi\big)-rV_\psi,
\]

where we set \(r=0\) for simplicity (it can be included). The PDE loss is then

\[
\mathcal{L}_{\text{PDE}}=\frac{1}{N_{\text{coll}}}\sum_{i=1}^{N_{\text{coll}}}|\mathcal{R}_{FK}(\mathbf{z}_i,t_i)|^2. \tag{7}
\]

Minimizing \(\mathcal{L}_{\text{PDE}}\) forces the latent dynamics to be consistent with the existence of a pricing measure that makes \(V_\psi\) a martingale. This is a soft constraint that encourages economic plausibility.

\subsubsection{Market Price of Risk Penalty}

Even if the PDE residual is small, the model might still produce implausibly high Sharpe ratios. To prevent this, we directly penalize the instantaneous Sharpe ratio. Define the market price of risk vector

\[
\boldsymbol{\lambda}(t)=\boldsymbol{\sigma}_\phi(\mathbf{z}(t),t)^{-1}\boldsymbol{\mu}_\theta(\mathbf{z}(t),t), \tag{8}
\]

assuming \(\boldsymbol{\sigma}_\phi\) is invertible (guaranteed by our parameterization). The squared norm \(\|\boldsymbol{\lambda}(t)\|^2\) represents the instantaneous Sharpe ratio (expected excess return per unit risk). To bound it, we introduce a hinge penalty:

\[
\mathcal{L}_{\text{MPR}}=\frac{1}{B}\sum_{b=1}^B\max\!\big(0,\|\boldsymbol{\lambda}(t_b)\|^2-\kappa^2\big), \tag{9}
\]

where \(\{t_b\}\) are sampled times and \(\kappa\) is a threshold. For daily data, a reasonable choice is \(\kappa=2\), corresponding to an annualized Sharpe ratio of about \(2\sqrt{252}\approx32\), which is already very high. This penalty discourages the model from learning strategies with unrealistic risk-adjusted returns.

\subsection{Forecasting and Consistency Objectives}

\subsubsection{Prediction Head}

The final prediction is obtained from the latent state at the horizon \(T\):

\[
\hat{y}=\mathbf{w}^\top\mathbf{z}_M+b, \tag{10}
\]

where \(\mathbf{z}_M\) is the SDE-simulated state at \(t_M=T\), \(\mathbf{w}\in\mathbb{R}^{d_z}\) and \(b\in\mathbb{R}\) are learnable parameters. The forecasting loss \(\mathcal{L}_{\text{forecast}}\) is the mean squared error for regression tasks or binary cross-entropy for classification.

\subsubsection{Consistency Loss}

To keep the latent trajectories grounded in the encoder outputs, we impose a consistency loss that penalizes deviations between the SDE-simulated states and the encoded states at each time step:

\[
\mathcal{L}_{\text{consist}}=\frac{1}{M}\sum_{j=1}^M\|\mathbf{z}_j^{\text{(sde)}}-\mathbf{z}_j^{\text{(enc)}}\|^2, \tag{11}
\]

where \(\mathbf{z}_j^{\text{(enc)}}=\mathcal{E}(\mathbf{x})(t_j)\) are the encoder outputs at the grid points. This acts as an auto-encoding regularizer, preventing the latent dynamics from drifting into unrealistic regions.

\subsection{Symbolic Bottleneck for Interpretability}

A key innovation of ARTEMIS is its ability to produce interpretable trading rules. We achieve this through a differentiable symbolic regression layer that distills the latent dynamics into closed-form expressions.

\subsubsection{Basis Function Library}

We predefine a library \(\mathcal{F}=\{f_1,\ldots,f_K\}\) of simple mathematical functions applied to the raw input features. Each \(f_k\) is a mapping from a window of length \(L\) of input features to a scalar. Typical functions include moving averages, ratios, differences, variances, and other elementary operations. For example, with a univariate time series \(\{x_t\}\),

\[
f_{\text{ma10}}(x)=\frac{1}{10}\sum_{i=1}^{10}x_i,\quad f_{\text{ratio}}(x)=\frac{x_1}{x_2},\quad f_{\text{diff}}(x)=x_1-x_2,\quad f_{\text{var}}(x)=\frac{1}{L-1}\sum_{i=1}^L(x_i-\bar{x})^2.
\]

In practice, we compute these functions for each feature channel and each possible lag, resulting in a large library.

\subsubsection{Sparse Linear Combination}

The symbolic layer forms a weighted combination of these basis functions:

\[
\hat{y}_{\text{symb}}=\sum_{k=1}^K w_k f_k(\mathbf{x}_{\text{input}}), \tag{12}
\]

where \(\mathbf{w}\in\mathbb{R}^K\) are learnable weights. To encourage sparsity and interpretability, we add an L1 penalty:

\[
\mathcal{L}_{\text{symb}}=\lambda_{\text{symb}}\|\mathbf{w}\|_1. \tag{13}
\]

\subsubsection{Differentiable Selection with Gumbel-Softmax}

For more flexibility, we can allow the basis functions themselves to have learnable parameters (e.g., the lag in a moving average). In that case, we use a Gumbel-Softmax relaxation to select among a set of candidate parameterizations. Let \(\alpha_k\) be logits for each candidate. The Gumbel-Softmax estimator provides a differentiable sample:

\[
p_k=\frac{\exp((\log\alpha_k+g_k)/\tau)}{\sum_{j=1}^K\exp((\log\alpha_j+g_j)/\tau)},\qquad g_k\sim\text{Gumbel}(0,1),
\]

where \(\tau>0\) is a temperature. The weighted combination becomes \(\hat{y}_{\text{symb}}=\sum_{k=1}^K p_k f_k(\mathbf{x}_{\text{input}})\). As \(\tau\to0\), this approximates a hard selection.

\subsubsection{Two-Phase Training}

To avoid interfering with the primary forecasting objective, we adopt a two-phase procedure:

\begin{enumerate}
    \item \textbf{Phase 1 (Pretraining):} Train the encoder, SDE, and prediction head using \(\mathcal{L}_{\text{total}}=\mathcal{L}_{\text{forecast}}+\lambda_1\mathcal{L}_{\text{PDE}}+\lambda_2\mathcal{L}_{\text{MPR}}+\lambda_3\mathcal{L}_{\text{consist}}\). The symbolic layer is not used.
    \item \textbf{Phase 2 (Distillation):} Freeze all parameters except those in the symbolic layer. Train the symbolic layer to mimic the full model's predictions using a teacher-student loss:
    \[
    \mathcal{L}_{\text{distill}}=\frac{1}{N_{\text{batch}}}\sum_{n=1}^{N_{\text{batch}}}(\hat{y}_{\text{symb},n}-\hat{y}_n)^2+\mathcal{L}_{\text{symb}}. \tag{14}
    \]
\end{enumerate}

This yields interpretable expressions that approximate the behavior of the full model.

\subsection{Conformal Prediction for Uncertainty Quantification}

To provide reliable uncertainty estimates, ARTEMIS incorporates conformal prediction, a distribution-free method that produces prediction intervals with finite-sample coverage guarantees.

\subsubsection{Standard Conformal Prediction}

Let \(\mathcal{D}_{\text{cal}}=\{(X_i,y_i)\}_{i=1}^n\) be a calibration set independent of the training data. For each calibration point, compute the absolute residual \(r_i=|y_i-\hat{y}(X_i)|\). For a new test point \(X_{\text{test}}\), construct the interval

\[
C(X_{\text{test}})=[\hat{y}(X_{\text{test}})-q_{1-\alpha},\,\hat{y}(X_{\text{test}})+q_{1-\alpha}], \tag{15}
\]

where \(q_{1-\alpha}\) is the \((1-\alpha)(1+1/n)\)-quantile of \(\{r_1,\ldots,r_n\}\). Under the assumption that the calibration and test points are exchangeable, we have the coverage guarantee

\[
\mathbb{P}(y_{\text{test}}\in C(X_{\text{test}}))\ge 1-\alpha. \tag{16}
\]

The proof follows from the fact that the ranks of the residuals are uniformly distributed (Vovk et al., 2005).

\subsubsection{Adaptive Conformal Prediction for Non-Stationary Data}

Financial time series are non-stationary, violating the exchangeability assumption. To address this, we employ an adaptive variant that maintains a rolling window of the most recent residuals. Let \(\mathcal{W}_t\) be a window of the last \(W\) residuals at time \(t\). The adaptive quantile \(q_{1-\alpha}(t)\) is the \((1-\alpha)\)-quantile of \(\mathcal{W}_t\). The prediction interval becomes

\[
C_t(X_{\text{test}})=[\hat{y}(X_{\text{test}})-q_{1-\alpha}(t),\,\hat{y}(X_{\text{test}})+q_{1-\alpha}(t)]. \tag{17}
\]

While this no longer provides a strict finite-sample guarantee, it adapts to changes in the error distribution and works well in practice (Gibbs \& Candès, 2021).

\subsubsection{Portfolio Optimization with Conformal Intervals}

The prediction intervals can be used for risk-aware portfolio construction. Consider a portfolio of \(P\) assets with weights \(\mathbf{w}\in\mathbb{R}^P\) satisfying \(\sum_{p=1}^P w_p=1\) and \(w_p\ge0\). For each asset, we have a point prediction \(\hat{y}_p\) and a conformal interval \([\hat{y}_p-q_p,\hat{y}_p+q_p]\). The continuous Kelly criterion maximizes the expected logarithmic growth rate:

\[
\max_{\mathbf{w}}\mathbb{E}[\log(1+\mathbf{w}^\top\mathbf{R})], \tag{18}
\]

where \(\mathbf{R}\) is the vector of returns. Using a quadratic approximation and the conformal intervals, we approximate

\[
\mathbb{E}[\log(1+\mathbf{w}^\top\mathbf{R})]\approx \mathbf{w}^\top\hat{\mathbf{y}}-\frac{1}{2}\mathbf{w}^\top\hat{\boldsymbol{\Sigma}}\mathbf{w}, \tag{19}
\]

where \(\hat{\boldsymbol{\Sigma}}\) is estimated from the conformal intervals (e.g., as a diagonal matrix with entries \(q_p^2\)). The optimization problem becomes

\[
\max_{\mathbf{w}}\mathbf{w}^\top\hat{\mathbf{y}}-\frac{\gamma}{2}\mathbf{w}^\top\hat{\boldsymbol{\Sigma}}\mathbf{w},\quad\text{s.t.}\ \mathbf{1}^\top\mathbf{w}=1,\ \mathbf{w}\ge0, \tag{20}
\]

with \(\gamma\) a risk aversion parameter. This convex quadratic program can be solved efficiently using differentiable convex optimization layers (Agrawal et al., 2019), enabling end-to-end training.

\subsection{Total Loss and Training}

The overall loss function combines all components:

\[
\mathcal{L}_{\text{total}}=\mathcal{L}_{\text{forecast}}+\lambda_1\mathcal{L}_{\text{PDE}}+\lambda_2\mathcal{L}_{\text{MPR}}+\lambda_3\mathcal{L}_{\text{consist}}+\lambda_4\mathcal{L}_{\text{symb}}. \tag{21}
\]

The hyperparameters \(\lambda_1,\ldots,\lambda_4\) balance the different objectives. Training proceeds by stochastic gradient descent. Gradients through the SDE solver are computed using the reparameterization trick: the Euler-Maruyama steps are deterministic functions of the initial state and the noise variables \(\{\boldsymbol{\epsilon}_j\}\), which are sampled independently of the parameters. Thus, we can backpropagate through the unrolled simulation using automatic differentiation.

\end{document}